\documentclass[submitting]{nst}

\usepackage{subfigure,dcolumn}
\usepackage[T2A,T1]{fontenc}
\usepackage[russian,english]{babel}
\usepackage{booktabs} 
\usepackage{graphicx} 

\usepackage{listings}
\begin{document}

\title{Artificial Intelligence in Reactor Physics: Current Status and Future Prospects}\thanks{Supported by the Natural Science Foundation of Shanghai (No. 23ZR1429300) and the Innovation Funds of CNNC (Lingchuang Fund, contract No. CNNC-LCKY-202234)}

\author{Ruizhi Zhang}
\affiliation{School of Mathematical Sciences, Shanghai Key Laboratory of Pure Mathematics and Mathematical Practice, East China Normal University, Shanghai 200241, China}
\author{Shengfeng Zhu}
\affiliation{School of Mathematical Sciences, Shanghai Key Laboratory of Pure Mathematics and Mathematical Practice, East China Normal University, Shanghai 200241, China}
\author{Kan Wang}
\affiliation{Department of Engineering Physics, Tsinghua University, Beijing 100084, China}
\author{Ding She}
\affiliation{Department of Engineering Physics, Tsinghua University, Beijing 100084, China}
\author{Jean-Philippe Argaud}
\affiliation{Électricité de France, R\&D, Palaiseau 91120, France}
\author{Bertrand Bouriquet}
\affiliation{Électricité de France, DQI 2 rue Ampère, Saint-Denis 93206, France}
\author{Qing Li}
\affiliation{Science and Technology on Reactor System Design Technology Laboratory, Nuclear Power Institute of China, Chengdu 610041, China}
\author{Helin Gong}
\email[Corresponding author. ]{gonghelin@sjtu.edu.cn.}
\affiliation{Paris Elite Institute of Technology, Shanghai Jiao Tong University, Shanghai 200240, China}

\begin{abstract}
Reactor physics is the study of neutron properties, focusing on using models to examine the interactions between neutrons and materials in nuclear reactors. Artificial intelligence (AI) has made significant contributions to reactor physics, e.g., in operational simulations, safety design, real-time monitoring, core management and maintenance. This paper presents a comprehensive review of AI approaches in reactor physics, especially considering the category of Machine Learning (ML, that we named also AI/ML to recall AI name we found in articles), with the aim of describing the application scenarios, frontier topics, unsolved challenges and future research directions. From equation solving and state parameter prediction to nuclear industry applications, this paper provides a step-by-step overview of ML methods applied to steady-state, transient and combustion problems. Most literature works achieve industry-demanded models by enhancing the efficiency of deterministic methods or correcting uncertainty methods, which leads to successful applications. However, research on ML methods in reactor physics is somewhat fragmented, and the ability to generalize models needs to be strengthened. Progress is still possible, especially in addressing theoretical challenges and enhancing industrial applications such as building surrogate models and digital twins.
\end{abstract}


\maketitle

\section{Introduction}
A nuclear reactor is a device for controlled nuclear reactions, i.e., a device that enables a self-sustaining nuclear chain reaction in a controlled manner. Depending on how atomic nuclei produce energy, nuclear reactors can be classified into fission reactors and fusion reactors. Currently, fission reactors are the most widely used in practical engineering applications.

Fission reactors have a wide range of applications, with electricity generation being the most common use, serving as the core component of a nuclear power plant. Typically, a fission reactor consists of nuclear fuel, coolant, moderator, neutron absorber, structural materials, and control mechanisms for regulating the fission process \cite{bib:1}.

\begin{figure*}
    \centering
    \includegraphics[width=0.8\linewidth]{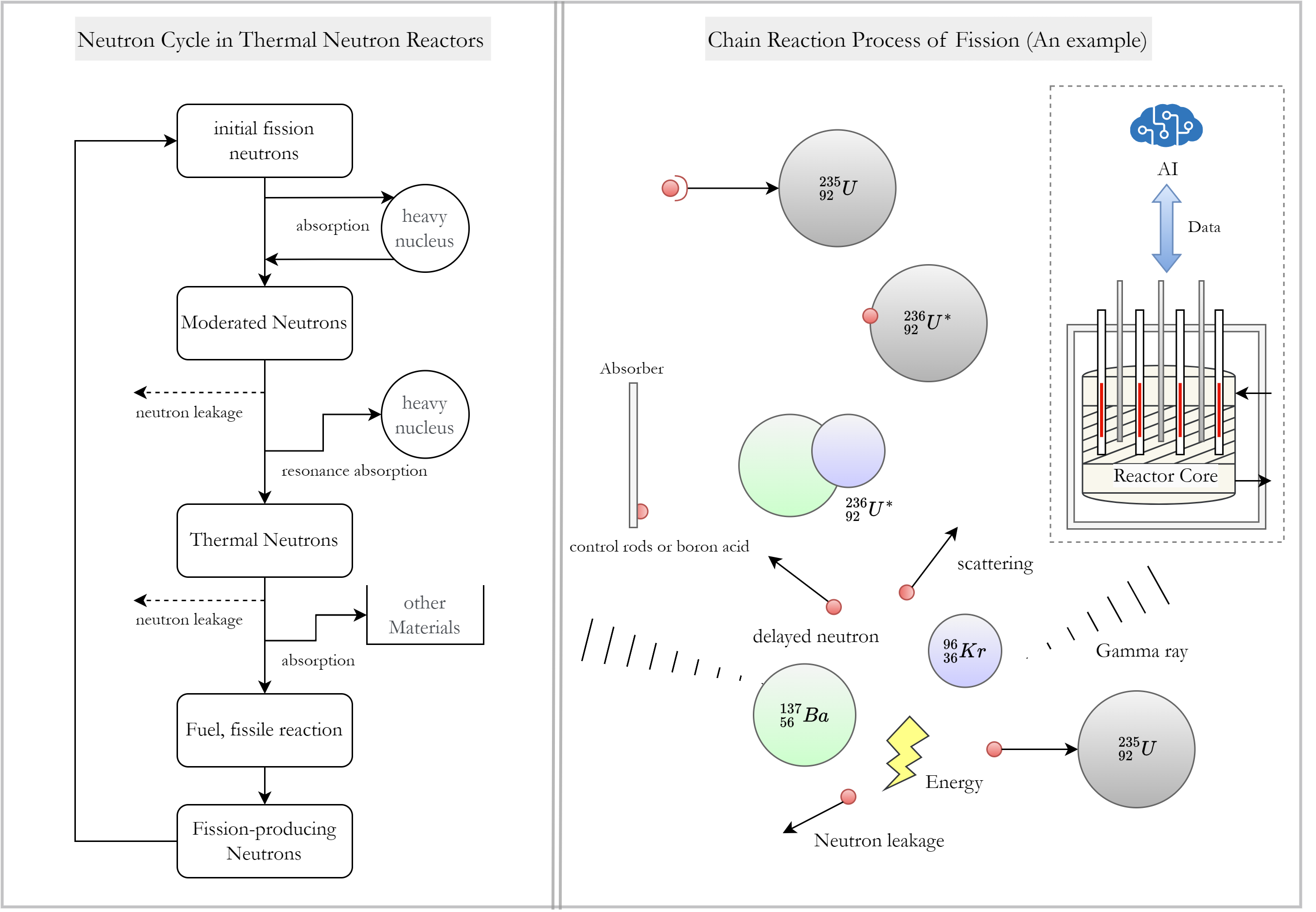}
    \caption{The process of a fission reaction in a nuclear reactor. A neutron interacts with a heavy nucleus (such as $\mathrm{^{235}U}$ or $\mathrm{^{239}Pu}$), triggering a fission reaction that splits the heavy nucleus into two or more smaller nuclei. This process releases a large amount of energy, mainly as the kinetic energy of the fission fragments, which is then converted into heat. During fission, additional neutrons are emitted, which can be absorbed, scattered, or trigger further fission events. If each fission-produced neutron successfully induces at least one new fission, the reaction becomes self-sustaining, forming a chain reaction. Control mechanisms within the reactor regulate this process by absorbing excess neutrons to maintain stability. The application of ML methods to reactor physics mainly considers the optimization of the parameters of this process, which are derived from the requirements of nuclear power plants.}
    \label{fig:1}
\end{figure*}

A fission reaction releases energy by splitting a heavy atomic nucleus into two (binary) or three (ternary) fragments \cite{bib:2}. Meanwhile, multiple new neutrons are produced, which can induce further fission reactions, thereby sustaining a self-sustaining chain reaction. The necessary condition for maintaining this chain reaction is that at least one of the newly created neutrons successfully causes a new fission event. However, not all neutrons contribute to further fission—some may be absorbed by non-fissile materials, while others may escape from the reactor core. The fission chain reaction is illustrated in Fig. \ref{fig:1}.

In general, there are many types of reactors, including light water reactors (LWR), heavy water reactors (HWR), gas-cooled reactors (GCR), fast breeder reactors\cite{bib:3} et. al. Among these, LWR is the most widely used and are further subdivided into boiling water reactors (BWR) and pressurized water reactors (PWR) \cite{bib:4}. In the case of LWR, for example, despite their proven safety and regulatory approval, the high construction costs and long-term operational challenges of large conventional LWRs have driven interest in developing fission reactors with significant improvements over the latest generation of fission reactors. These include LWR designs that are much smaller than existing reactors, and concepts that use different moderators, coolants, and fuel types \cite{bib:4}. The main types of advanced reactors include advanced water-cooled reactors (WCR), GCR, liquid metal-cooled reactors (LMR), molten salt reactors (MSR), and fusion reactors, and any research aimed at optimizing reactors requires the analysis of reactor properties or the prediction of parameters.

Nowadays, most of the literature may not directly refer to the study of "reactor physics", but refers to the terms "neutronics", "nuclear reactor theory" or "reactor analysis", which have almost similar meanings \cite{bib:5}, with the commonality that the relevant methods or research ideas are used, or that the physics related to the neutrons of the fission reaction is involved, which is also the method of definition in this review. Frequent research objectives in the literature are the study of the determination of the neutron position in reactor cores, the reaction of neutrons with the surrounding medium, and the study of the spatio-temporal distribution of neutrons, including the improvement of efficiency by ML methods. In addition, the synthesis and organization of the literature related to neutronic interactions with matter in nuclear reactors by AI/ML methods is still fragmented. For numerical simulation techniques in the field of reactor physics, the latest publication \cite{bib:6} classified the studies into six categories: (i) nuclear data processing and resonance calculation models, (ii) cross-section homogenization techniques, (iii) steady-state and transient neutron transport methods, (iv) Monte Carlo approaches and applications, (v) nuclear reactor design and analysis, and (vi) methods for sensitivity and uncertainty analysis. Inspired by this research, we will investigate the field of reactor physics, collate the works of literature focusing on frontier applications of methods in ML to the field, and demonstrate what reactor physics is, how it can be solved with the help of ML methods to solve problems related to it, and how to enhance and improve the practical efficiency of the methods.

\subsection{What the Reactor Physics is}

A nuclear power plant is a complex system, and the production of nuclear energy must be operated under very high safety standards \cite{bib:7}. Therefore, predicting the distribution of neutrons throughout the reactor is both critical and challenging. As one of the main reactants in a nuclear reactor, neutrons will interact with various materials in the reactor core. 

Reactor physics is essentially the physics of neutron properties in a reactor core \cite{bib:6}. It focuses on methods that use analytical and numerical models to study the interaction between neutrons and matter in reactor cores, which is the basis for reactor design and analysis \cite{bib:6}. The method can be used to solve a range of problems in the simulation and regulation of reactor cores in nuclear power plants. Reactor physics models construct high-fidelity datasets with the help of deterministic and indeterministic methods to obtain critical neutron information of reactor cores \cite{bib:6}, to realize the process of operation simulation and design of real reactors on computers during operation and to realize the maintenance of the nuclear power plant system as well as the safety supervision \cite{bib:9}. The ML-based reactor physics model is an extension of the neutron transport model, which aims to provide more accurate or more efficient predictions of neutronic behaviors under certain (maybe more complex) material or geometric configurations, where many-queries simulations or real-time computations are required, for design or optimal operation purposes \cite{bib:10}.

Back to 1956, Weinberg \cite{bib:11} mentioned in his book on nuclear reactor physics that the theory was mainly concerned with the neutron distribution in nuclear reactors, and mentioned that the central goal of reactor technology was to generate useful work from nuclear reactors. This provided the basis and ideas for later research on nuclear reactor physics. During the same period, more books appeared around the general direction of the goals of reactor physics \cite{bib:12, bib:13, bib:14}. In 2005, Zin \cite{bib:1} organized the foundations and advances in nuclear reactor physics and integrated the two main goals of reactor physics. They focused on the design and analytical methods of this period, problems and limitations, and related research to address them, and gave seven mainstream directions for research, which we list as follows \cite{bib:1}.
\begin{itemize}
    \item Derivation of the multigroup transport equations and the multigroup diffusion equations, with representative solution methods thereof.
    \item Elements of modern (now almost three decades old) diffusion nodal methods.
    \item Limitations of nodal methods such as transverse integration, flux reconstruction, and analysis of UO2-MOX mixed cores. Homogenization and related issues.
    \item Description of the analytic function expansion nodal (AFEN) method.
    \item Ongoing efforts for three-dimensional whole-core heterogeneous transport calculations and acceleration methods.
    \item Elements of spatial kinetics calculation methods and coupled neutronics and thermal-hydraulics transient analysis.
    \item Identification of future research and development areas in advanced reactors and Generation-IV reactors, in very high-temperature gas reactor (VHTR) cores.
\end{itemize}
It is worth pointing out that Marguet's book \cite{bib:15} on the challenges facing reactor physics, the recommendations of which are compiled below, could provide some thoughts for our next efforts within AI framework.

\begin{itemize}
    \item Resolution of a 3D core with fine deterministic transport, in "one go" killing the two-step procedure.
    \item Depletion, or kinetic, Monte Carlo calculations applied to exact geometries.
    \item On-line codes continuously fed by experimental measurements.
    \item Better operation of the instrumentation of present reactors or future instrumentation.
    \item New computational schemes substituting the two-group diffusion method with multi-group methods using simplified transport.
    \item More prevalent multi-physics coupling of specialized codes so that the physical modeling of phenomena becomes consistent, which means not having high-performing physics in one field being associated with too simplistic models in another.
    \item Adjoint codes with high perturbation orders to finally obtain uncertainties not solely based on the renowned but controversial "engineer's judgment".
\end{itemize}

\subsection{What the Artificial Intelligence Methods are in Literature}

In recent years, ML methods, particularly neural networks (NNs) and their variants, have made significant contributions to the advancement of nuclear reactor physics. However, it is crucial to recognize that the application of AI/ML in this field is not limited to NNs alone. Other static learning methods, such as support vector machines (SVM), random forests (RF), and support vector regression (SVR), also play pivotal roles in addressing various challenges in reactor physics.

Whether in the simulation of nuclear reactor operation, the design of the reactor core, or the real-time monitoring, management, and maintenance of nuclear power plants, AI/ML methods have been widely applied. The current advanced methods are already present in the industrial applications of nuclear power plants, but still suffer from the problem of mutual constraints of accuracy and time scale. As powerful autonomous learners, the emergence of AI/ML methods, including but not limited to NNs, precisely presents a better solution to these problems.

\begin{figure}
    \centering
    \includegraphics[width=1\linewidth]{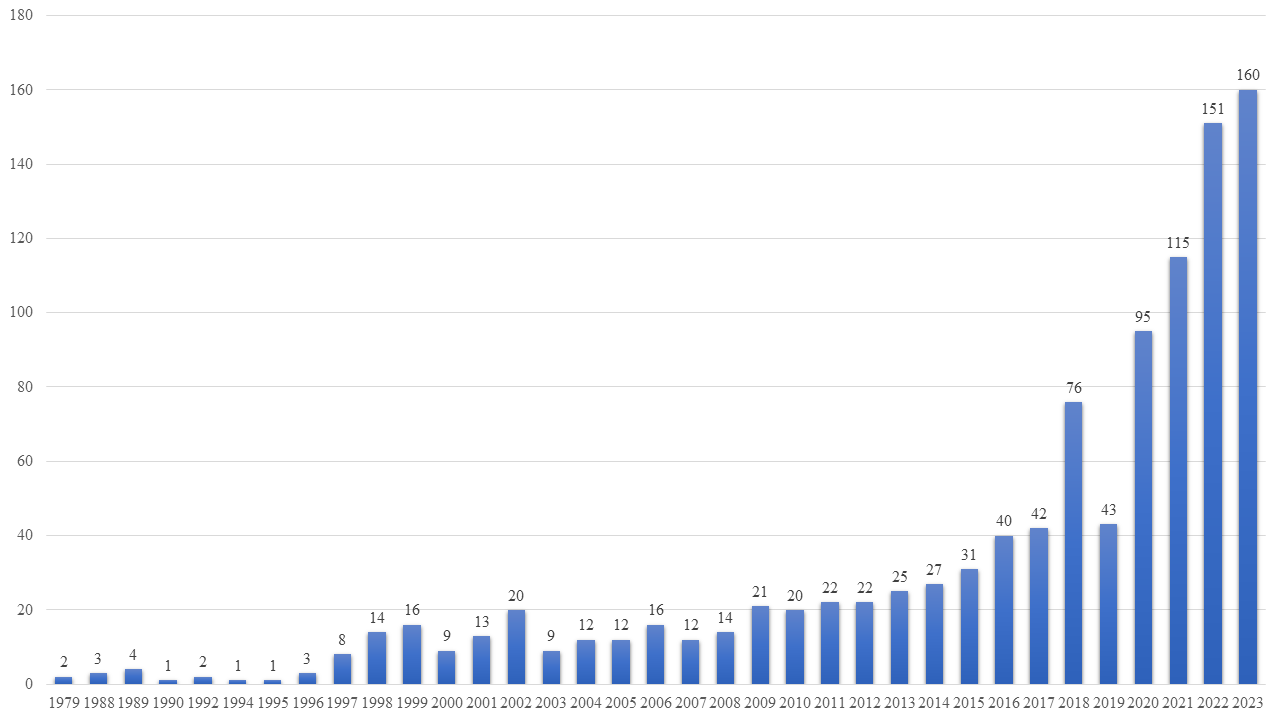}
    \caption{ Literature Publication Trends in Reactor Physics: This histogram illustrates the annual publication count from the earliest available data through 2023, highlighting a notable surge in research output since 2020. The horizontal axis denotes the years, and the vertical axis shows the number of publications, offering insights into the growing academic interest in this field.}
    \label{fig:2}
\end{figure}

\noindent
\textit{1) The comparatively early research on ML in nuclear reactor physics can be traced back to the end of last century.}

These literatures involve simple ML methods and focus on scalar aspects of prediction, such as power, temperature, etc. The application scenarios are relatively limited mainly due to the immaturity of computer technology and neural network technology. The application of ML methods in reactor physics has not been developed a lot. The literature showing the relevant typical cases is as follows. Kim et al. \cite{bib:16} investigated the fuel reloading problem in a pressurized water reactor (PWR), i.e., the two major constraints of keeping the local power peaking factor below a predetermined value and maximizing the effective multiplication factor ($k_\text{eff}$) for the efficient operation of the reactor. They developed a fast core parameter prediction model based on backpropagation neural network (BPNN) \cite{bib:16}. 

\noindent
\textit{2) With the recent advances in reactor physics \cite{bib:1}, the first decade of the 21st century has a gradual increase in research on the application of ML methods to reactor physics. We list some typical literature from this period.}

Studies built neural networks to improve the application. Ortiz et al. \cite{bib:17} used neural networks to predict the core parameters of BWR. They further used a polymorphic recurrent neural network (RNN) to optimize the fuel loading pattern of the BWR \cite{bib:18}, which was used as a basis for designing a fuel lattice optimization system for the BWR \cite{bib:19}. Kozlowski et al. \cite{bib:20} used Artificial Neural Network (ANN) to develop an innovative approach for the practical application of the Pin-Cell Discontinuity Factors (PDFs). 

The Support Vector Regression (SVR) method has brought some improvements in the field of applications. For instance, based on this method, Bae et al. \cite{bib:21} discussed its application in calculating the reactor power peaking factor (PPF), while Trontl et al. \cite{bib:22} explored its application in optimizing the reactor loading pattern.

\noindent
\textit{3) By the second decade of this century, especially in 2018, the application of ML methods in reactor physics started to gradually explode. }

During this period, with the further development of reactor physics techniques, especially digital simulation techniques, coupled with the maturity of computer technology, the application of ML methods to reactor physics has become progressively more sophisticated and can deal with some more complex application scenarios. 

Based on the Hopfield neural network artificial (HNNA), Tayefi et al. \cite{bib:23} designed a new method to guide the heuristic search and distribute the axial variation of enrichment to make the neutron flux smoother. Based on multilayer perceptron (MLP), Pirouzmand et al. \cite{bib:24} implemented the prediction of key neutron parameters in the core and constructed a real-time monitoring system. Fernandez et al. \cite{bib:25} evaluated the ability to predict the behavior of the system during accidents with power input and flux loss in the core. Based on cellular neural networks, Starkov et al. \cite{bib:26} constructed a prediction system for heavy water moderator temperature. Based on the SVR method, Zeng et al. \cite{bib:27} constructed a system performance prediction model consisting of a reactor physical model and a thermo-hydraulic model based on a Transportable Fluoride-salt-cooled High-temperature Reactor (TFHR).

\noindent
\textit{4)	From 2020 to the present, there is a lot of literature focusing on the application of ML methods to the field of nuclear reactor physics. }

ML techniques, especially neural network techniques, are rapidly developing. There has been a surge in research in the field of ML-based approaches to reactor physics with highly sophisticated application scenarios. Based on a combination of wavelet analysis and convolutional neural network (CNN), Tagaris et al. \cite{bib:28} proposed a new technique for detecting signal anomalies in nuclear reactors.

Also advanced is the field of prediction of fundamental parameters or phenomena of reactor physics. For instance, Bamidele et al. \cite{bib:29} predicted the decay heat of LWR fuel assemblies based on the Gaussian process (GP), support vector machine (SVM) model, and neural network (NN). Berry et al. \cite{bib:30} trained ANN and random forest (RF) classifiers to determine whether the given energy structure enables a multi-group collision probability model to calculate accurate neutron multiplication factors in LWR lattice simulations. Based on deep neural network (DNN), Chen et al. \cite{bib:31} investigated a linear transport model, while Alam et al. \cite{bib:32} developed a multi-stage prediction model. Based on ANN, Dorde et al. \cite{bib:33} on the other hand, studied a Sodium Fast Reactor (SFR), where the model efficiently evaluates the possible Doppler reactivity of the reactor core over its lifetime range, and Xie et al. \cite{bib:34} solved the neutron diffusion problem for continuous neutron flux distribution. Moreover, Chen et al. \cite{bib:35} screened the burnup state and predicted neutron parameters in a liquid fuel MSR with the help of the LightGBM model.

Faced with complex application scenarios, Sobes et al. \cite{bib:36} developed an ML-based algorithm for the design and optimization of a model for the shape of the nuclear reactor core. Turkmen et al. \cite{bib:37} used ML in the single-channel design of MSR. Meanwhile, based on the application scenarios, digital twin monitoring techniques have also been developed. Gong et al. \cite{bib:10, bib:38} proposed a digital twin combining a reduced-order model with data-enabled physics-informed machine learning (ML) to realize the monitoring of power, where the k-nearest neighbor (KNN) algorithm and a decision tree algorithm were involved to predict the operating parameters and enable online monitoring of the power distribution. Furthermore, Prantikos et al. \cite{bib:39} improved the digital twin for nuclear reactor monitoring. 

Benefiting from the proposal and development of physics-informed neural network (PINN) techniques, theoretical problems in reactor physics have been able to propose efficient solutions. In 2019, Raissi et al. \cite{bib:40} proposed PINN, which has been widely used in nuclear reactor physics in the recent past. With the help of the PINN method, the literature presents learning methods for solving linear transport equations \cite{bib:41}, new techniques for the numerical solution of neutron diffusion equations \cite{bib:42}, methods for solving point kinetic equations (PKEs) \cite{bib:39}, and solving neutron diffusion equations for single-energy and multi-energy groups \cite{bib:43}. 

New application scenarios based on reactor physics have given rise to new NN structures, which are a further step up from the simple NN ones of the previous phase. The main research here is the improvement of the PINN algorithm for specific problems and the derivation of many variants. These include deep jointly-informed neural network (DJINN) for the prediction of fission and scattering products \cite{bib:44}, conservative PINN (cPINN) for solving heterogeneous neutron diffusion problems with non-smooth solution \cite{bib:45}, boundary dependent physics-informed neural network (BDPINN) for solving neutron transport equations \cite{bib:46}, and physically-informed neural network with migration learning (TL-PINN) for reactor transient prediction \cite{bib:47}. There are also more complex variants along with applications. Yang et al. \cite{bib:48} investigated generalized inverse power method neural network (GIPMNN) and physically constrained GIPMNN (PC-GIPMNN) methods to determine eigenvalues. To further improve the accuracy and efficiency, they also proposed a so-called data-enabled PINN (DEPINN) \cite{bib:7}.

\begin{figure}
    \centering
    \includegraphics[width=1\linewidth]{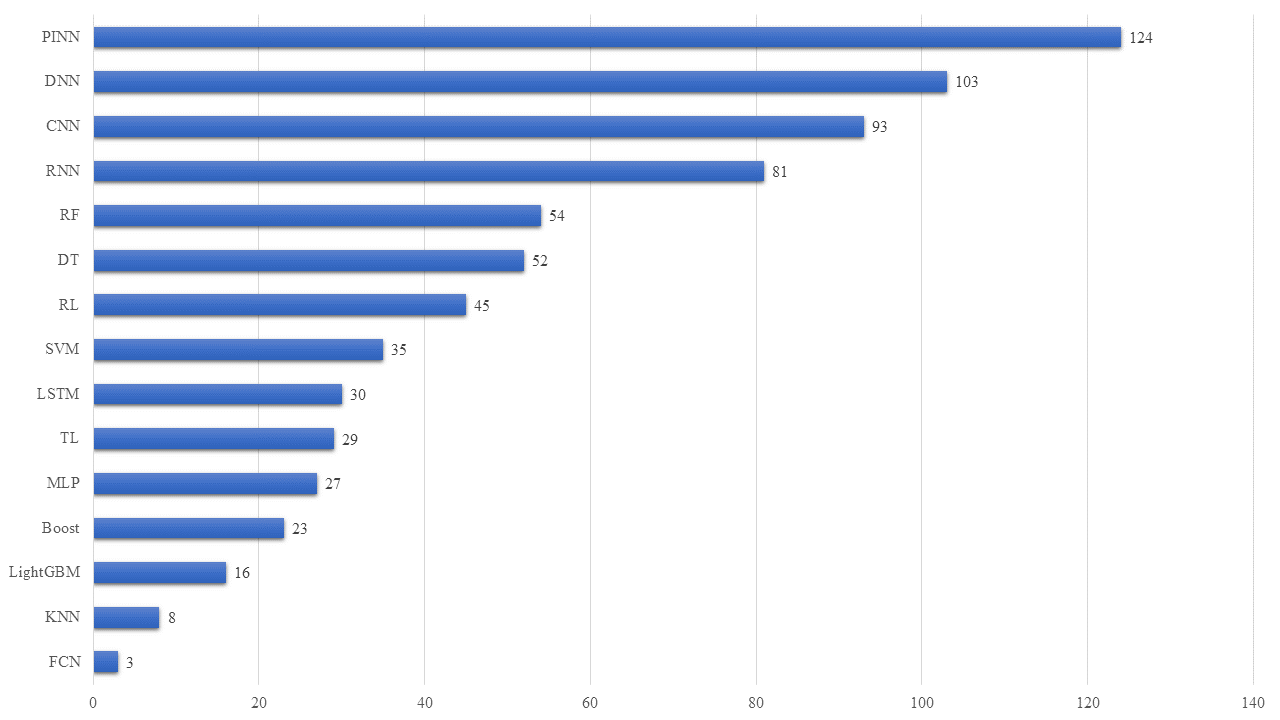}
    \caption{Application of AI/ML Methods in Reactor Physics: A Comparative Analysis. This bar chart categorizes and compares the frequency of various AI/ML methodologies in the literature, including CNN, DNN, RNN, Reinforcement Learnin (RL), Transfer Learning (TL), Long and Short-Term Memor (LSTM), and Fully Convolutional Networks (FCN) within Deep Learning (DL), as well as the specialized PINN \cite{bib:49} approach. The x-axis lists the ML methods, and the y-axis indicates the corresponding number of publications, reflecting the diverse algorithmic applications in reactor physics research.}
    \label{fig:3}
\end{figure}

There are many types of AI methods applied in reactor physics or its related fields: KNN, decision tree, multilayer perceptron (MLP), boosting algorithm, ANN, PINN, etc. With the help of Scopus platform search, we collected the frequency of occurrence of these methods, which is now demonstrated in Fig. \ref{fig:3}.

\begin{figure}
    \centering
    \includegraphics[width=1\linewidth]{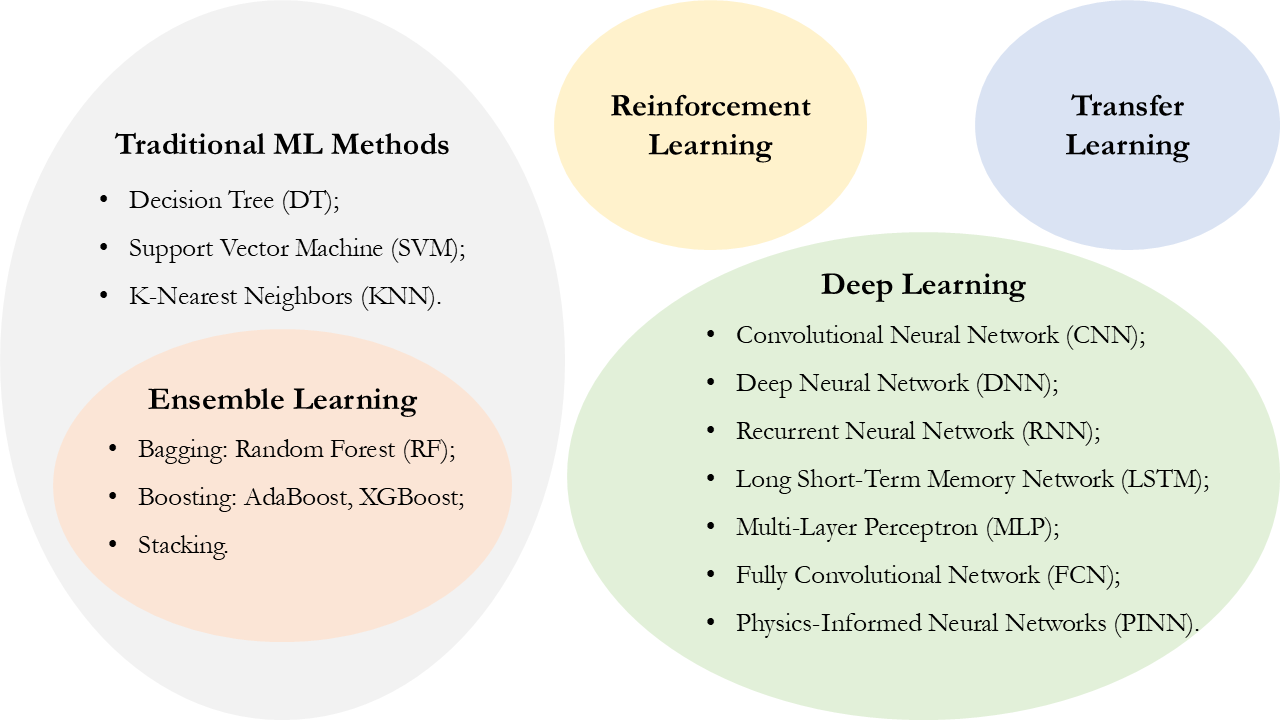}
    \caption{In this paper, we delineate the taxonomy and interplay of ML methods. Traditional methods, particularly those grounded in ensemble learning, frequently serve as the foundation for predicting core parameters in reactor physics research. Deep learning stands out as the most extensively studied and cutting-edge approach, with a multitude of methodological variants emerging to address specific challenges in the field. Reinforcement learning and transfer learning, as burgeoning disciplines, are poised for further exploration within the domain of reactor physics. These areas hold promise for enhancing the efficiency of learning from data and for optimizing model transferability to mitigate the time costs associated with training. Future research is likely to delve into these methods, investigating their potential to streamline the predictive capabilities and computational efficiency of reactor physics models.}
    \label{fig:4}
\end{figure}

Current ML methods in reactor physics range from traditional methods such as integrated learning to NN-based deep learning, reinforcement learning, and transfer learning. Fig. \ref{fig:4} shows the categorization and connections of these methods for easy reading.

In summary, the application of ML methods in literature related to reactor physics analysis has gone far beyond the basic applications including reactor core design, simulation, and safety regulation. It involves the mathematical mechanisms behind them, including the solution of the neutron transport equation, neutron diffusion equation, point kinetic equations, etc., also for different scenarios realizing the industrial demands. Higher accuracy in predicting important reactor parameters will contribute to better fuel planning and compliance with technical specifications, while increased efficiency in prediction will bring advances in the management of nuclear power plants, especially in the regulation and maintenance of nuclear reactors. Therefore, for the frontiers of reactor physics, the application scenarios are unfolding from steady state to transient, the former of which focuses on core simulation and design, while the latter will bring the necessity of nuclear fuel safety regulation.

Through the application of ML methods, reactor physics to solve industrial demands has been able to grow at a high rate. The advantages brought about by the application of ML to reactor physics are, firstly, the realization of efficiency gains in simulation and design due to the ability of ML to learn on its own, which directly improves its ability to adapt to engineering problems. The next is that it overcomes the real-time problem in the traditional framework. This is due to ML's driving research on theories in reactor physics, especially some important mathematical theories. The application extends beyond the fission reaction process to encompass all aspects of nuclear reactors, including nuclear data processing \cite{bib:50}, thermo-hydraulics \cite{bib:27,bib:51}, chemical kinetics \cite{bib:52}, radiation transport problems \cite{bib:53}, etc. For example, ML methods can be utilized to calculate wall temperature parameters under certain conditions \cite{bib:54}, identify and resolve potential biases in data \cite{bib:55, bib:56}, and data reduction \cite{bib:10, bib:44}. Furthermore, there will be applications to solve problems related to nuclear fusion \cite{bib:57, bib:58}. Indeed, the development of reactor physics has been unstoppable with ML.

\subsection{What is this Review About}

Actual operation of the reactor core is best illustrated by the distribution of power, or by the observation of the neutron flux, while the multiplication factor directly determines whether the chain reaction is carried out or not \cite{bib:1}. Considering the safety design and optimal operation of a nuclear reactor, it is essential to determine important properties such as temperature, neutron flux, reaction rate or power distribution, cross-section, and fuel burnup in reactor cores, including both global state and local details. The analysis of the reactor should involve the determination and prediction of the state of the reactor that sustains the chain reaction by balancing neutrons from fission and losses from capture and leakage, all based on reactor physics \cite{bib:3}. More specifically, the goal of reactor physics is to determine neutron multiplication factor for various configurations of a reactor, respectively, and neutron flux distributions, spatial and temporal, under various operating conditions \cite{bib:1}.

Based on the analysis above, in this survey, we focus on how AI/ML can be used to address different problems of reactor physics, e.g., solving the neutron transport equation, the building blocks of reactor physics, aspects related to transport theory, the toolset that is already available, future directions and recent trends, as well as issues related to sensitivity and uncertainty analysis. Depending on these problems, we show how high-fidelity nuclear reactor simulations can be customized in literature through parameter prediction, data augmentation, and the use of transfer learning.

\begin{figure*}
    \centering
    \includegraphics[width=0.9\linewidth]{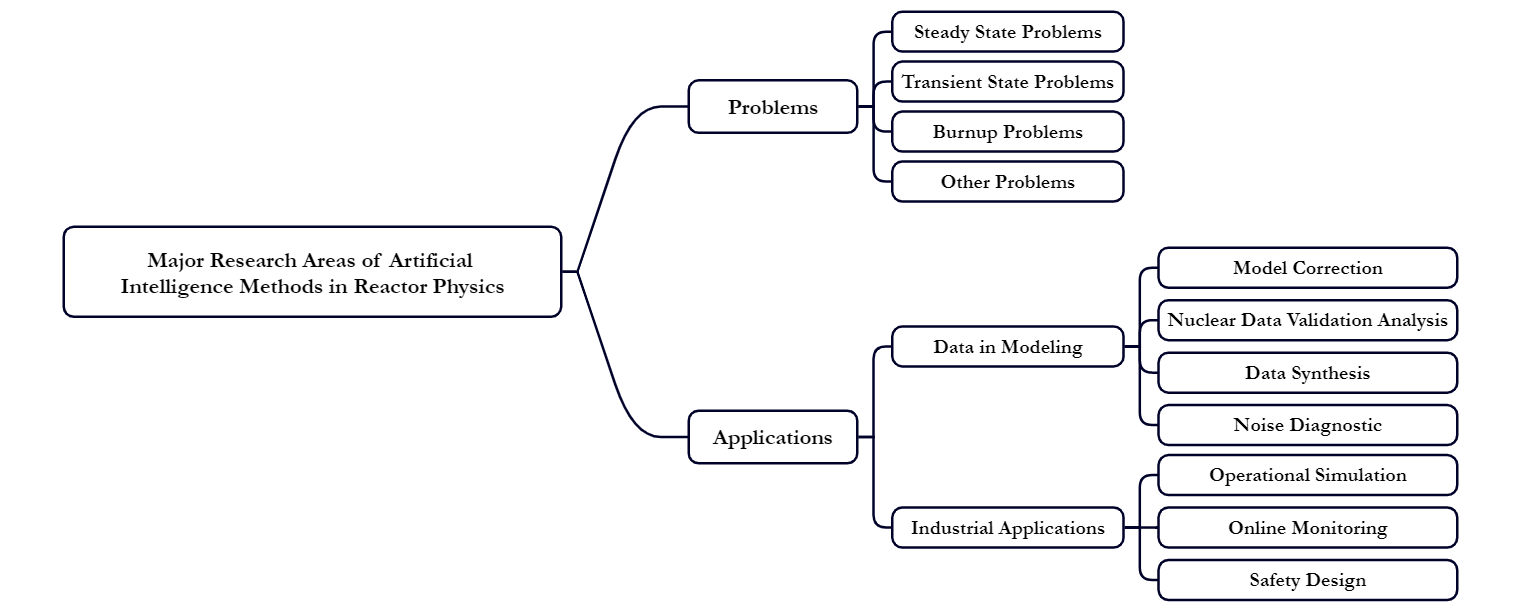}
    \caption{The mind map shows practical challenges and application scenarios of ML methods in the field of reactor physics. On one hand, this paper takes a cut through some of the challenges that need to be solved urgently in industry, including the steady state and transient problems derived from the neutron spacetime variables, and the burnup problems containing the fuel where the nuclei of the atoms to be reacted are located, which have been facilitated by ML to achieve good effects; on the other hand, there are several industrial applications based on these problems, including advanced data processing methods, and the practical impetus brought by ML in the operational simulation, online monitoring, and safety design.}
    \label{fig:5}
\end{figure*}

This review aims to cover most of the frontier ML-related articles related to nuclear reactor physics, and to learn about advances in machine learning in nuclear physics, refer to the review \cite{bib:69a}. A practical scenario of the ML approach to reactor physics in the article is shown in Fig. \ref{fig:5}. We stress that the objective of this work is not to compare the performance of ML methods since they were developed for different goals and to address different problems. Thus, this paper can serve as a comprehensive guide for navigating these fast-growing ML-related techniques and methodologies in reactor physics. The primary research question is to determine what reactor physics is and the ML-based applications to the simulation, design, and safety regulation of nuclear reactors. In addition, the review considers researches that address a broader topic origin from but more than reactor physics, which includes important theories of nuclear reactors such as thermal hydraulics. 

\section{Fundamentals of Reactor Physics Theory}

Theoretical analysis of reactor physics can address issues such as neutron behavior in fission reactors and the interaction of neutrons with individual materials. Regarding neutron transport processes, the governing equations portray them. In the standard description of neutron transport, the neutron flux is related to the position, flight direction, energy, and time of the neutron. The development of traditional numerical methods is mainly driven by the high demands on accuracy and time in the industrial simulation. 

In multiscale problems, energy is typically divided into different hierarchical levels. If neutrons in the reactor core are assumed to have isotropic properties within a specific reaction zone, the directional component of neutron flight can be neglected. This leads to the simplification of the model into two categories: steady-state and transient equations, which are further elaborated in Section 3 of this paper. The steady-state approach focuses on reactor behavior over extended periods, examining the reactor's lifetime and its final state, without taking individual neutron reactions into account. This removes the correlation between the neutron flux and time, as governed by the neutron diffusion equation. On the other hand, transient equations, such as the point kinetic equations (PKEs), focus on the time evolution of the system. In this case, the spatial position of individual neutrons is irrelevant; instead, the neutron population and the reaction rates are modeled as a function of time, describing the overall kinetics of the reactor rather than the movement of individual neutrons.

While the previous considerations focused on the neutron behavior in the reaction, the material composition of the reactor core also significantly affects the reaction dynamics. Burnup is described by a set of differential equations that model the consumption and production of nuclear fuel, commonly referred to as burnup equations. Burnup is crucial for the study of nuclear reactors, including aspects such as fuel reloading and the power peaking factor. Proper management of burnup helps to ensure neutron flux equilibrium and extends the lifetime of the nuclear reactor. Additionally, the burnup distribution plays a crucial role in analyzing core transients. Specifically, for short-duration transients, the burnup of the core is assumed to be constant, as the depletion process occurs over a much longer timescale compared to the duration of transients, which can range from seconds to days.

Especially since the development of ML technology, data can drive the model. From an engineering perspective, ML approaches can directly predict neutron flux distributions, enhancing reactor simulations without traditional discretization methods or expressed by parameters such as the multiplication factor and the reaction cross-section. From this point of view, parameter prediction in nuclear reactors has also been more extensively studied. In this aspect, some studies focus directly on power prediction, which is intuitively an important parameter in reactor control.

\begin{figure*}
    \centering
    \includegraphics[width=0.9\linewidth]{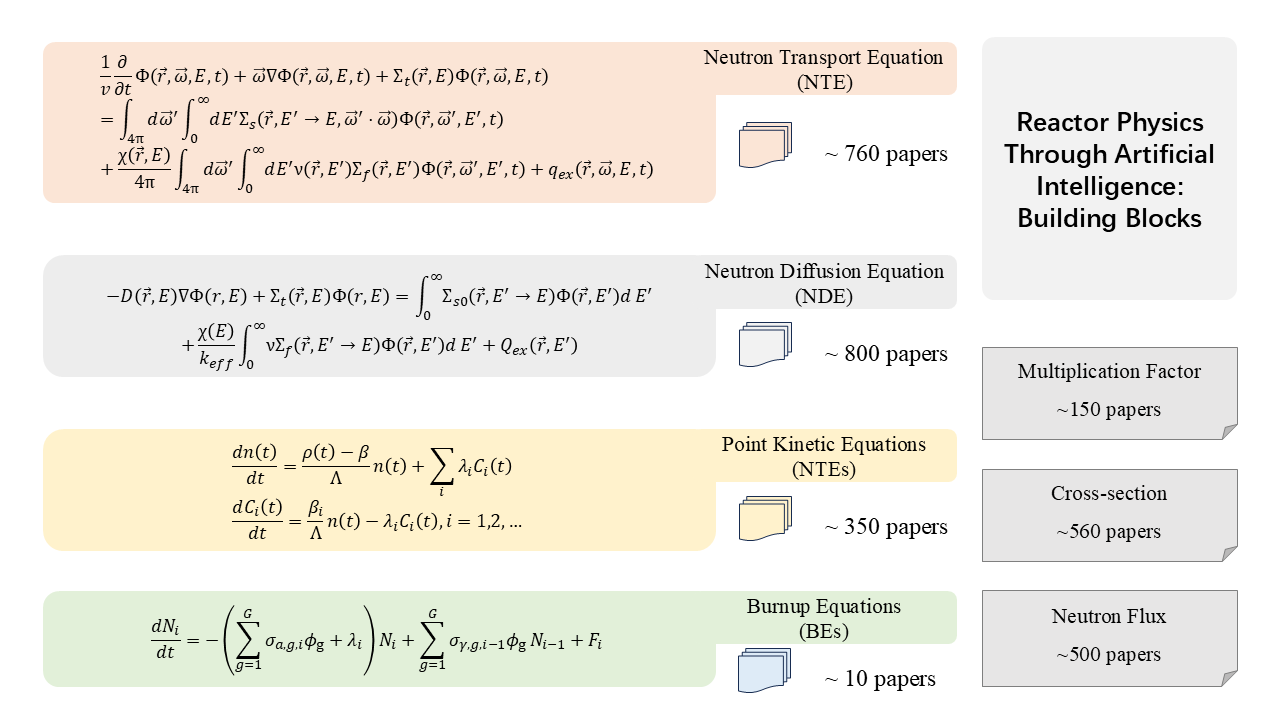}
    \caption{Utilizing search data from the Scopus platform, we conducted a literature review focusing on the foundational elements of reactor physics. Our analysis revealed a significant body of work that integrates ML methods to address complex challenges within the field. By examining the core structure of reactor physics, which encompasses the governing equations of nuclear reactions and pivotal state parameters, we have quantified the prevalence of scholarly references in recent years, facilitated by the Scopus platform. There is a discernible surge in research dedicated to the intricate equations governing neutron transport, neutron diffusion, and point kinetics, as well as a pronounced trend towards the study of critical parameters that are essential for reactor physics.}
    \label{fig:6}
\end{figure*}

Fig. \ref{fig:6} summarizes the fundamental components of reactor physics, distinguishing between the governing equations and key model/engineering parameters. The early research in reactor physics focused on the deterministic approach, i.e., the solution process controlled by the four equations mentioned above, which is the first block focused on ML applications in reactor physics. The second segment is the direct or indirect prediction of some important parameters. Their predictions go far beyond their value and are more of a catalyst for solving other problems.

\subsection{Governing Equations}
Regarding the mathematical model describing the neutron transport phenomenon, the Boltzmann transport equation \cite{bib:59}, which is one of the key equations in nuclear engineering and reactor physics, is commonly used. It is an integral-differential equation that describes the neutron flux \cite{bib:60} across the cross-section of an oriented plane in an inhomogeneous fissile medium, and is an accurate description of the actual neutron transport process in the medium. To be specific, neutron interactions with the medium in the process of transport will occur, such as scattering and absorption reactions, and these will affect the neutron flux. Neutron transport equations are based on the principles of conservation of both energy and momentum, and by taking these interactions into account, the behavior of neutrons in a reactor or other medium is predicted. The Boltzmann transport equation takes the following form 
\begin{equation}
\begin{aligned}
    \frac{1}{v} \frac{\partial \Phi}{\partial t}
    + \vec{\omega} \cdot \nabla \Phi
    &+ \Sigma_{t} \Phi = 
    \int_{4 \pi} d \vec{\omega}' \int_{0}^{\infty} d E' \, \Sigma_{s} \, \Phi' \\
    & + \frac{\chi}{4 \pi} \int_{4 \pi} d \vec{\omega}' \int_{0}^{\infty} d E' \, v \, \Sigma_{f} \, \Phi'+ q_{ex}.
\end{aligned}
\label{eq:1}
\end{equation}

In equation (\ref{eq:1}), $\Phi(\vec{r}, \vec{\omega}, E, t)$ denotes the neutron flux, which depends on the position vector $\vec{r}$, energy $E$, and direction of motion $\vec{\omega}$ at time $t$. This flux is a fundamental parameter in nuclear reactor physics, as it describes the distribution and intensity of neutrons in the reactor. The term $\Phi'$ refers to the neutron angular flux at energy $E'$ and direction $\vec{\omega}'$, representing contributions from neutrons scattering into the specified state $(\vec{r}, \vec{\omega}, E, t)$ from other energies and directions. The neutron transport equation, in its most general form, involves seven independent variables: three spatial coordinates in $\vec{r}$, two angular coordinates in $\vec{\omega}$, energy $E$, and time $t$ \cite{bib:9}. This comprehensive model, though accurate, is often simplified in practical reactor design through approximations such as the diffusion equation.

Equation (\ref{eq:1}) can describe the neutron transport behavior in a medium precisely. More specifically, in the transport process, neutrons will interact with the medium and undergo reactions such as scattering and absorption, and these will affect the neutron flux. In the equation, $\Sigma$ denotes the macroscopic cross-section, i.e., the probability that a certain process occurs for a neutron, for example, collision, which includes scattering (either elastic and inelastic scattering), and  absorption (either capture, or fission, or others). Other notions are standard.

Neutron transport theory models neutron behavior with greater precision by incorporating angular dependence, particularly important in highly detailed reactor simulations. In contrast, the diffusion approximation simplifies this by assuming isotropic neutron motion. In contrast, the neutron diffusion equation is considered a simplification of the neutron transport equation and is commonly used in core calculations for nuclear reactor design and analysis. The theory is considered accurate in relatively homogeneous and large geometries because neutrons tend to move isotropically, i.e., without directional preference \cite{bib:13}. The neutron diffusion equation is generally of the following form:
\begin{equation}
-D \nabla \Phi + \Sigma_{t} \Phi = \int_{0}^{\infty} \Sigma_{s 0} \, \Phi' \, d E' +\frac{\chi}{k_{\text{eff}}} \int_{0}^{\infty} v \Sigma_{f} \, \Phi' \, dE' + Q_{ex}.
\label{eq:2}
\end{equation}

While neutron diffusion equation have been successfully employed in reactors of various sizes, their application can lead to significant errors due to the underlying diffusion theory. These errors arise from several factors, including the diffusion approximation itself, spatially integral representations, and simplifications in energy cross-section interactions. In particular, regions with large gradients of neutron current (flux), especially near reactor boundaries, can amplify these errors. Such spatial approximations and their associated inaccuracies are well-established features of the method, rather than unforeseen limitations\cite{bib:14, P1}. There are several numerical methods for solving this equations, including finite element methods (FEM), finite volume methods (FVM), and finite difference methods (FDM) \cite{bib:34}. However, the accuracy of these methods is essentially limited by the number of nodes and geometry of the problem \cite{bib:61}.

As the basis of diffusion theory, Fick's law is essentially related to the $P_1$ approximation of the time-dependent neutron transport equation (the approximation is also known as the neutron telegraph equation), and the diffusion equation further neglects the derivative term of the neutron flux density with respect to time. Since the parabolic nature of the diffusion equation predicts that the particles will have infinite velocities, the particles in the tails of the distribution function will appear at infinite distance from the source in time $t = 0+$, but the neutron telegraph equation has finite particle velocities, the value of which is incorrectly \cite{P1}. Thus in some sense it cannot accurately characterize the actual neutron transport process\cite{bib:62}.

On the other hand, for the case where the spatial dependence of the neutron flux is negligible, reduced-order models consisting of PKEs have been developed \cite{bib:39}. PKEs consist of a system of coupled nonlinear stiff differential equations that model the kinetics of reactor variables. This is a reduced-order model of the Boltzmann neutron transport equation and the Bateman equation describing the 3D spatial and temporal kinetics of a nuclear reactor \cite{bib:47}. The PKEs are derived under the approximation of neglecting the shape of the neutron flux and the neutron density distribution \cite{bib:47}. 

PKEs, first deduced by Henry \cite{bib:63}, reduce the mathematical problem from a system of integral differential equations to a system of ordinary differential equations \cite{bib:64}, which allows for the rapid solution of time-dependent neutron transport problems. The equation is generally written in 0D and has the following basic form \cite{bib:65}.
\begin{equation}\label{eq:4}
    \begin{aligned}
        \frac{dn}{dt} &= \frac{\rho - \beta}{\Lambda} n + \sum_i \lambda_i C_i,\quad i = 1, 2,\ldots \\
        \frac{dC_i}{dt} &= \frac{\beta_i}{\Lambda} n - \lambda_i C_i, \quad i = 1, 2, \ldots
    \end{aligned}
\end{equation}
In equation (\ref{eq:4}), $n$ stands for the neutron density concentration, $C_i$ is the delayed neutron precursor density concentration for group $i$, $\rho$ is the reactivity feedback, which are functions of time $t$, $\beta_i$ is the delayed neutron fraction for each group, and $\beta = \sum_i \beta_i$
is the sum of the delayed neutron fractions. In addition, $\Lambda$ is the mean neutron lifetime in the reactor core, and $\lambda_i$ is the average decay constant of the $i$-th group precursor. Given the appropriate initial conditions, this constitutes a complete set of ordinary differential equations that reflect information about the power level and power fluctuations of a nuclear reactor in a unit of time \cite{bib:65}.

In this paper, we plan to integrate the extented mainstream ML techniques for solutions to important governing equations in reactor physics. Table \ref{tab:1} contains a synthesis of the kind of AI/ML approaches that have been synthesized in the field of reactor physics, as well as the literature in which the approach has been used.

Recently, with the development of ML technology, the use of ML, especially deep learning (DL), to solve complex differential equations is becoming a hot research topic in the field of numerical computation \cite{bib:66}. The advantage of resorting to DL instead of traditional neural networks is that it can solve the problem of high computational cost for problems with high-dimensional physical variables and can handle complex boundary conditions \cite{bib:40}. Based on artificial neural networks (ANNs), in literature, DL methods have been proposed to solve the neutron diffusion problem for continuous neutron flux distributions without the need for prior regional discretization \cite{bib:34}.

In the process of a neutron reaction, the fissile isotopes in the fuel are depleted continuously due to neutron-induced reactions and radioactive decay, and will also change with the reaction \cite{bib:67}. To calculate the changes in the composition within the fuel during reactor operation accurately, it is necessary to establish a function of the change in each of the isotopes involved in the reaction over time, that is, the burnup equation. The burnup equation is generally written in the form of equation (\ref{eq4}).
\begin{equation}
    \frac{dN_i}{dt} = - \left( \sum_{g=1}^{G} \sigma_{a,g,i} \Phi_g + \lambda_i \right) N_i + \sum_{g=1}^{G} \sigma_{\gamma,g,i-1} \Phi_g N_{i-1} + F_i,
    \label{eq4}
\end{equation}
where $N_i$ denotes the number of the $i$-th nuclide, and $\sigma_{a,g,i}$, $\sigma_{\gamma,g,i}$ denote the absorption cross-section and fission cross-section of the $i$-th nuclide. In the formula, the right-hand side contains the disappearance and production of nuclides, respectively, where the first term denotes the disappearance rate of the isotope $i$ due to absorption of neutrons and decay neutrons, the second term denotes the production rate of the isotope $i$ due to absorption of neutrons or decay of the isotope $i - 1$, and the third term denotes the production rate due to the fission reaction \cite{bib:68}.

It is worth noting that, in general, the neutron angular flux $\Phi$ is a function of both spatial location $r$ and time $t$, and that there is an interaction between the $N_i$ of the fuel isotope and the neutron angular flux $\Phi$. Thus, strictly speaking, equation \eqref{eq4} is a nonlinear problem. In practical operations, it is common to divide the core into several burnup regions in which the position does not vary much, and thus the variables are independent of $r$. Furthermore, given the burnup time step, the neutron flux can be approximated as constant, thus creating a system of ordinary differential equations for each burnup step with respect to time $t$.

Among others, convolutional neural networks (CNNs) \cite{bib:69} and deep neural networks (DNNs) \cite{bib:31} have been applied to various problems in reactor physics. However, one of the hottest research topics in deep learning (DL) is the physics-informed neural network (PINN). Unlike traditional data-driven machine learning (ML) methods—which require large training datasets—PINNs can directly solve higher-order, multi-dimensional equations in a forward manner \cite{bib:66}. The extensive literature on PINNs \cite{bib:41,bib:42,bib:43,bib:53} highlights their growing importance in addressing complex physical problems.

Physics-informed neural networks (PINNs) incorporate the governing equations (e.g., the neutron diffusion equation~\eqref{eq:2} or the Boltzmann transport equation~\eqref{eq:1}) directly into the neural network loss function. Specifically, the PDE residual and any boundary or initial conditions are enforced during training, allowing the network to learn solutions that inherently respect the underlying physics. By leveraging automatic differentiation (AD) to compute derivatives with respect to both the network parameters and input variables, PINNs offer a mesh-free approach that can handle complex reactor geometries and high-dimensional parameter spaces. Variants such as conservative PINN (cPINN) \cite{bib:45}, physically constrained generalized inverse power method neural network (PC-GIPMNN) \cite{bib:48}, and boundary-dependent PINN (BDPINN) \cite{bib:46} further extend this framework to address domain decomposition, boundary constraints, or improved training stability. These methods provide a promising pathway for coupling data-driven models with fundamental reactor physics to enhance both accuracy and computational efficiency. 

\subsubsection{Neutron Transport Equation}
The neutron transport equation delineates the distribution of neutron flux across a cross-sectional plane within a heterogeneous fission medium \cite{bib:60}. In principle, this equation is capable of yielding more precise solutions. Nonetheless, the high-fidelity data obtained from solving the transport equation comes at the cost of extensive computational time and substantial computational resource demands. These limitations often render it impractical for iterative core design in commercial applications. Conversely, the merits and demerits of grid-based methods are increasingly acknowledged in real-world applications. The accuracy of these methods is fundamentally constrained by the granularity of the grid — quantified by the number of nodes — and the geometric complexity of the problem domain \cite{bib:70}.

\begin{table*}[!ht]
    \centering
    \caption{The main ML methods used in the governing equations of the reaction.}
    \renewcommand{\arraystretch}{1} 
    \vspace{0.5em} 
    \begin{tabular}{p{1.5cm} p{3.5cm} p{9cm} p{3.1cm}} 
        \toprule
        NN Family & NN Type & Solving Problems & Papers \\ 
        \midrule
        NN & Celluar NN & Spatial and temporal response of the neutron flux distribution & Pirouzmand et al. \cite{bib:71} \\ 
        ANN & ANN & Albedo problem, the Milne problem, and the criticality problem & Türeci \cite{bib:74} \\ 
        DNN & DNN & Solution of the neutron transport equations & Liu et al. \cite{bib:73}, \newline Chen et al. \cite{bib:31} \\ 
        PINN & PINN & Loosely Coupled Reactor Model (LCRM) & Elhareef et al. \cite{bib:75} \\ 
        & PINN, DEPINN, GIPMNN & Solution of neutron diffusion equations \newline Solution of neutron transport equations & Huhn et al. \cite{bib:53}, \newline Yang et al. \cite{bib:7}, \newline Yang et al. \cite{bib:76}, \newline Yang et al. \cite{bib:48} \\ 
        & PINN & Solution of point kinetic equations & Prantikos \cite{bib:39}, \newline Schiassi et al. \cite{bib:64} \\ 
        & Boundary-Dependent PINN & Solution of the neutron transport equations & Xie et al. \cite{bib:46} \\ 
        & Conservative PINN & Heterogeneous seed diffusion problems with non-smooth solutions & Wang et al. \cite{bib:45} \\ 
        & TL-PINN & Reactor Transients (RTs) Prediction & Konstantinos et al. \cite{bib:77} \\ 
        \midrule
        \bottomrule
    \end{tabular}
    \label{tab:1}
    \vspace{0.5em} 
    \parbox{\textwidth}{\footnotesize \textit{We provide a synthesis of the literature on the application of machine learning (ML) methods to the governing equations describing neutron behavior in nuclear reactors. Among these, the Physics-Informed Neural Network (PINN) approach has emerged as a particularly prominent technique, extensively applied and refined in recent studies. As a deep learning-based framework, PINNs have demonstrated significant efficacy in obtaining numerical solutions to the governing equations, establishing themselves as a robust and innovative tool for advancing reactor physics research.}}
    \vspace{1em} 
\end{table*}

\noindent\textit{Cellular Neural Network}

Cellular neural network is one of the alternatives to traditional numerical methods \cite{bib:71}. Based on the cellular neural network, Pirouzmand et al. \cite{bib:71} simulated the spatial-temporal response of the neutron flux distribution under steady state and transient conditions with the help of the second-order form of the time-dependent neutron transport equation and simulate step perturbations in the core of the reactor. They also invoked the model, used on a small pressurized water reactor assembly to simulate the effects of temperature feedbacks, poisons, and control rods on the neutron flux distribution.

\noindent\textit{Deep Neural Network}

Chen et al. \cite{bib:31} refer to the traditional $S_N$ method \cite{bib:72}, which discretizes the angle variable of the product of the definite integral term, constructs the approximate discretization of the definite integral term and forms the loss function by Gaussian product group, and gives the results of discrete-time and angle scalar flux density calculation of one-dimensional geometry in the transient case by DNN.

Furthermore, Liu et al. \cite{bib:73} propose a differential transform order theory for the difficulty of DNN in solving the neutron transport equation with a definite integral term, which first converts the transport equation with a calculus form into a higher-order differential equation, and then uses DL based on NN to approximate the original function, and performs a differential downscaling of the original function to obtain the numerical solution of the angular flux density of the transport equation.

\noindent\textit{Physics-informed Neural Network}

Based on DL frameworks, PINN-based methods also show great advantages compared to traditional methods. Huhn et al. \cite{bib:53} solved the stream and interaction terms of the Boltzmann transport equation by PINN, proposed the application of Fourier features and a heuristic-based sampling method to improve the performance of PINN, and verified the advantages of the algorithm in terms of accuracy with the example of a transmission problem in a heterogeneous one-dimensional plate. To eliminate the error caused by boundary conditions and avoid the curse of dimensionality, Xie et al. \cite{bib:46} proposed a boundary-dependent physically informed neural network (BDPINN) for solving the neutron transport equation. They also introduced three techniques to improve BDPINN, including third-order tensor transformation, rearranging the training set, and resulting in reconstruction in higher order, which optimizes the computational cost and accuracy.

\subsubsection{Neutron Diffusion Equation}

Solutions derived from the neutron diffusion equation are extendable to a variety of industrial applications, including the simulation and design of nuclear reactors. Advanced simulations of reactors often incorporate multi-physics modeling and feature intricate geometries, which demand more sophisticated analytical capabilities than those provided by conventional grid-based methods.

\noindent\textit{Artificial Neural Network}

Türeci \cite{bib:74} leveraged established single-velocity neutron transport data with isotropic scattering to address the albedo, Milne, and criticality problems using two distinct ML methods: polynomial regression (PR) and artificial neural networks (ANN). The PR method demonstrated efficacy within the confines of its training data scope, while the ANN method, despite requiring more computational time, exhibited robustness beyond the training data limits.

\noindent\textit{Deep Neural Network}

Dong et al. \cite{bib:42} constructed a DNN as a trial function and substituted it into the neutron diffusion equation to form the residuals, which are used as the weighted loss function of the PINN, so as to approximate the numerical solution of the diffusion equation by deep machine learning. They also proposed techniques such as accelerated convergence methods, efficient parallel search techniques for the proliferation factor, and strategies for learning the inhomogeneous distribution of the sample network points, and analyzed the sensitivity of each key parameter of the neural network.

\noindent\textit{Physics-Informed Neural Network}

Elhareef et al. \cite{bib:75} applied a PINN to solve a loosely coupled reactor model (LCRM) based on the neutron diffusion model. They used a two-dimensional, time-domain independent, constant source diffusion equation to simulate the LCRM problem with zero inflow flux physics conditions, which were reduced to general Robin boundary conditions. Relative to the FEM solution, PINN obtained an average relative error of about 0.63\%.

For the problem of neutron diffusion with heterogeneous neutrons, Xie et al. \cite{bib:34} proposed two mesh-free physical information deep learning (PIDL) methods, including the boundary-dependent method (BDM) and the boundary-independent method (BIM), which give continuous symbolic solutions to solve the neutron diffusion problem with continuous neutron flux distribution. This procedure does not require region discretization and thus can be easily generalized to complex geometries. In tests dealing with complex geometries and multi-region problems, the BDM method shows higher accuracy and flexibility. Research \cite{bib:78} also solved the coupled system of partial differential equations by the PINN method, respectively, and solved the multi-energy group diffusion problem based on a system of equations consisting of several diffusion equations with neutron fluxes and $K$ eigenvalues, while Wang et al. \cite{bib:45} introduced the concept of cPINN, where PINN is developed for each subdomain and additional conservation laws along the subdomain interfaces are taken into account, and the method is proved to be effective in solving the heterogeneous neutron diffusion problem with non-smooth solutions. More PINN-related works will be discussed in detail in Section 3.

\subsubsection{Point Kinetic Equations}
The point reactor kinetics (PRK) model is derived from the point kinetic equations (PKEs). This model assumes that the flux shape function follows the fundamental mode, which is the solution to the external neutron source-free balance equation. This approximation remains valid primarily when the system is near criticality, where fission sources dominate over external sources \cite{bib:79}.

Compared to traditional numerical methods, the PINN method does not require the time-consuming construction of complex grids and has a short computational time, so it can be applied more efficiently to solve problems on irregular and high-dimensional domains. Considering the advantages of PINN, Prantikos et al. \cite{bib:39} developed a PINN model for solving PKEs and obtained results that are consistently compared to standard numerical methods. Schiassi et al. \cite{bib:64} proposed a new method for solving temperature feedback PKEs based on the PINN method, which is flexible and problem-adaptable. Gatchalian et al. \cite{bib:80} explored the role of the ML method in utilizing measurable quantities from reactor operations and experiments as predictors of $k_{\text{eff}}$ and kinetic/subcritical parameters.

Transfer learning physics-informed neural network (TL-PINN), the process of pre-training a neural network on similar data to enhance performance on a new task, is currently less used in transient reactor physics problems. Konstantinos et al. \cite{bib:77} achieved an improvement in reactor transient (RTs) prediction performance with the help of a TL-PINN method, which reduces the number of model iterations. Specifically, they used a PRK model with six neutron precursor populations, constructed using experimental parameters of the Purdue University Reactor-1 (PUR-1) research reactor, to generate different reactor RTs with a range of experimentally relevant variables, while the similarity was characterized using Hausdorff and Fréchet distances. The model yields accelerations of up to two orders of magnitude. The average error in the prediction of neutron densities by the conventional PINN and TL-PINN models is less than 1\%.

\subsection{Key Reactor Parameters}

In reactor physics, the state parameters are predominantly characterized by the neutron transport equation, its simplified counterpart, the neutron diffusion equation, or the point kinetic equation. These equations can be addressed through both deterministic approaches and probabilistic methods, such as the Monte Carlo method. Regardless of the chosen method, the overarching objective is to accurately model key state parameters within a nuclear reactor. These include the multiplication factor, neutron cross-sections, and neutron flux, which are the principal manifestations of the fission reaction—a process that is microscopically intricate and complex. Table \ref{tab:2} provides an overview of the ML techniques employed in the prediction of these parameters, as reported in existing literature.

\begin{table*}[!ht]
    \centering
    \caption{ML methods for the prediction of state parameters important in the reaction.}
    \renewcommand{\arraystretch}{1} 
    \setlength{\tabcolsep}{12pt} 
    \begin{tabular}{lll}
        \toprule
        State Parameters & ML methods & Papers \\ 
        \hline
        Multiplication factor & ANN, RF & Berry et al. \cite{bib:30} \\ 
        & BPN & Kim et al. \cite{bib:81}, Bei et al. \cite{bib:82} \\ 
        & LSTM & Ren et al. \cite{bib:83} \\ 
        & Multi-state RNN & Ortiz et al. \cite{bib:19} \\ 
        & SVR & Trontl et al. \cite{bib:22} \\ 
        Neutron cross-section & DNN, XGBoost & Oktavian et al. \cite{bib:70}, Sun et al. \cite{bib:84} \\ 
        & DNN & Ravichandran et al. \cite{bib:85} \\ 
        & Decision Tree, KNN & Vicente-Valdez et al. \cite{bib:86}, Li et al. \cite{bib:87} \\ 
        Neutron flux & ANN & Xie et al. \cite{bib:34} \\ 
        & CNN & Berry et al. \cite{bib:88} \\ 
        & FCN & Zhang et al. \cite{bib:69} \\ 
        & HNNA & Sadighi et al. \cite{bib:89} \\ 
        & KNN, Decision Tree & Gong et al. \cite{bib:10} \\ 
        \midrule
        \bottomrule
    \end{tabular}
    \label{tab:2}
    \vspace{0.5em} 
    \parbox{\textwidth}{\footnotesize \textit{We highlight an additional significant domain where ML methods are applied within the framework of reactor physics: the prediction of state parameters. The methodologies for parameter prediction are more varied than those employed for solving the governing equations of the nuclear reactions. This diversity underscores the field's ongoing potential for research innovation. It is noteworthy that this area often intersects with the preceding section, given that the equations in question encapsulate the microscopic dynamics of neutron transport—a fundamental and substantial foundation for the accurate prediction of state parameters.}}
    \vspace{1em} 
\end{table*}

\subsubsection{Multiplication Factor}

The multiplication factor as an important quantity in reactor physics is a measurement of the change in the number of fission neutrons from one neutron generation to the next, and is a direct reflection of the ability of a nuclear reactor to maintain conditions for a self-sustaining chain reaction. It is vital to achieve an accurate prediction of the multiplication factor, which determines the critical conditions of the reactor, and accurate values that enable reactor operators to make informed decisions about fuel management and overall safety.

The original application of ML methods focused on predicting the multiplication factor, and at the same time, the PPF. Kim et al. \cite{bib:81} developed a fast core parameter prediction system based on a pressurized water reactor (PWR) using back propagation neural networks (BPNs) to model the multiplication factor and the maximum power, which resulted in a large speed improvement with errors within a few percent. Ortiz et al. \cite{bib:19} introduced a method based on a multi-state recurrent neural network to optimize the nuclear fuel lattice system of a BWR to predict the local PPF and the infinite multiplication factor ($k_{\text{inf}}$). Trontl et al. \cite{bib:22} used the support vector regression (SVR) algorithm for the global $k_{\text{eff}}$ at the cycle begin and end, respectively, as well as the PPF as the target parameter, respectively, to build an ML model and optimize the parameters by genetic algorithms (GA), and obtained a good result with the root mean square error (RMSE) controlled in the order of $10^{-2}$.

The application of ML methods in reactor physics is maturing and has achieved better results, especially on accuracy improvement in recent studies. Berry et al. \cite{bib:30} trained an ANN and random forest (RF) classifier to determine whether a given 20-group energy structure enabled a multi-group collision probability model to compute accurate neutron multiplication factor in LWR lattice simulations, and the trained model achieved classification with 95.3\% and 95.5\% accuracy, respectively. Ren et al. \cite{bib:83} explored the potential of utilizing Long Short-Term Memory (LSTM) neural networks, to predict the multiplication factor in nuclear reactor physics, and modeled the multiplication factor with the first cycle of loading the BEAVRS core for 0-300 days of full-power operation, and adjusted the appropriate parameters to make the predicted $k_{\text{eff}}$ to within 2 pcm absolute error. 

Oktavian et al. \cite{bib:70} first developed a DNN-based error correction model with high-fidelity data for a 2*2 BWR corelset, generated by the Monte Carlo method. They utilized multiple ML methods to generate high-fidelity data from the error correction of low-fidelity data, and compared to traditional numerical methods like Monte Carlo simulations, the results of DNN and XGBoost methods are better in the results, the error of $k_{\text{eff}}$ for PARCS assisted with DNN and XGBoost correction models can reduce the error from 200-300 pcm to around 50 pcm or lower on average. Bei et al. \cite{bib:82} developed a BP neural network-based surrogate model for fast prediction of multiplication factor in the core, based on data modeled and simulated through the use of a Monte Carlo method for the different operating states of the MSR experiment. The maximum absolute error (MAE) of the $k_{eff}$ prediction surrogate model is reduced to approximately 70 pcm, significantly improving the accuracy of core calculations compared to traditional physics-based models, where most errors are within ± 50 pcm, with a maximum absolute error of 70 pcm, demonstrating the reliability of the keff surrogate model.

\subsubsection{Neutron Cross-section}

In terms of the two-step approach which is currently the most used in nuclear reactor simulations, the first step is the generation of uniform few-group cross-sections \cite{bib:90}. Taking high-temperature gas-cooled reactors (HTGR) as an example, to improve the accuracy of the reactor physics module of the HTGR Engineering Simulation System (HTR-ESS), more accurate computational models of the cross-section are generally built in real-time simulation. However, due to the complex nonlinear relationship between the cross-section parameters and other state parameters, there is no generalized model \cite{bib:86, bib:91}.

Ravichandran et al. \cite{bib:85} trained DNN to predict the value of the cross-section of 2D pin cell model and 2D lattice model nodes of a PWR and confirmed the feasibility of this method in PWR. They also mentioned the optimization of the algorithm with the help of transfer learning (TL) to reduce the dependence on the big dataset.

Vicente-Valdez et al. \cite{bib:86} proposed to support complex tasks of nuclear reactor evaluators with the help of ML technology. They developed decision tree and KNN methods to fit nuclear data and infer the induced reaction cross-section for neutrons. The predicted values based on ML methods matched the new measurements quite accurately. In their work, they explored proof-of-concept models that rely on learning the underlying patterns of cross-section data from other radionuclides, demonstrating that ML models can help traditional physics-guided models and play a role in nuclear data evaluation. They \cite{bib:86} also pointed out that incorporating the ML model into the nuclear data module can enable evaluators to make faster, unbiased decisions in areas of uncertainty and better inform data measurement activities in the most sensitive areas of future Experimental Nuclear Reaction Data (EXFOR). Finally, Ivanov \cite{bib:92} explored the use of ML methods to address the problems faced when approximating neutron cross-sections using, in particular, NNs.

For industrial applications, especially in the absence of data, Sun et al. \cite{bib:84} investigated the $(n,2n)$ reaction cross-section of fission products by ANN and XGBoost models to predict these reaction cross-sections. ANN predictions with MAPE deviations less than 10\% accounted for more than 85\% of the test set. To meet the real-time requirements in ESS, Li et al. \cite{bib:87} proposed the use of DNN to consider the complex nonlinear relationships among the reactor variables and used tree regression (TR) to obtain the accurate cross-section and optimize the cross-section calculation among them.

For different problems, ML methods need to be used with caution. Martin et al. \cite{bib:90} investigated a DL framework for the mathematical representation of multi-group cross-sections in the Griffin reactor multi-physics code for two-step deterministic neutronic calculations. They provided a technique for propagating cross-section model errors to $k_{\text{eff}}$ using sensitivity coefficients with first-order uncertainty propagation rules. They also showed that MLP achieves accuracy from 1765 to 800 pcm in cross-sections with nonlinear dependence, especially in the case of rough networks. For linearly dependent multigroup cross-sections in sodium-cooled fast reactors (SFR), simple linear regression will be better.

The cross-section prediction will be beneficial for high-precision fuel depletion calculations. Leniau et al. \cite{bib:93} discussed the characterization of the spatial distribution of Pu in nuclear fuel as well as the establishment and validation of burnup models. They obtained the spatial distribution of Pu with the help of the Latin Hypercube Sampling (LHS) method and proposed a method based on NNs, which can be used to build a fuel loading model for PWR loaded with MOX fuel and a cross-section predictor, and the average cross-section predictor can complete the fuel consumption calculations with high accuracy in a short time.

Related studies have also explored the internal structure of the cross-section in terms of its internal structure. To determine the resonant structure of the cross sections, Nobre et al. \cite{bib:94} applied ML methods to automatically assign the angular momentum quantum numbers of the resonances.

\subsubsection{Neutron Flux}

ML techniques have been used in neutron flux flattening problems as early as 2002, when Sadighi et al. \cite{bib:89} proposed to combine a continuous Hopfield NN with a simulated annealing algorithm (SA) to guarantee an optimal solution with reduced computational effort. In solving the neutron flux flattening problem in Bushehr nuclear power plant a good result was obtained.

To address the problem of calculating the continuous neutron flux distribution, Xie et al. \cite{bib:34} proposed two boundary processing methods based on ANN and compared their accuracy and efficiency by analyzing benchmark tests. The experimental results show that the latter is more accurate. They further proposed the trial function construction method to generalize its application.

Zhang et al. \cite{bib:69} proposed a DL-based agent model to replace the traditional diffusion equation solver and thus predict the flux and power distributions of reactor cores. They designed a CNN inspired by a fully convolutional network (FCN) and obtained positive results.

Combining reduced-order modeling and ML methods, Gong et al. \cite{bib:10} proposed a data-enabled physics-informed machine learning (PIML) digital twin model to predict high-dimensional outputs such as neutron flux distributions and power distributions in reactor cores. With the help of KNN and decision tree, they attempted to show the effectiveness of HPR-1000 for online monitoring. The article also explores how the model can be used to quickly reconstruct the neutron field for real-time input parameters.

Furthermore, Bei et al. \cite{bib:82} developed a BPNN-based proxy model for fast prediction of the channel-by-channel neutron flux distribution in the 3D channel of an MSR core. They generated dataset samples by simulating different operating states of the MSR using Monte Carlo methods and trained and optimized the proxy model. Also in MSR, Chen et al. \cite{bib:35} analyzed the neutron flux with the help of multiple regression methods. As mentioned previously, Elhareef et al. \cite{bib:43} applied the PINN method to a neutron diffusion model to solve a loosely coupled reactor model and used it to optimize the NN hyper-parameters, which resulted in a significant reduction in the predicted neutron flux error.

ML techniques have also been used for data enhancement, with Berry et al. \cite{bib:88} exploring the use of CNN to improve the resolution of neutron flux statistics in Monte Carlo simulations.

\section{Problems in Reactor Physics Dealt with ML Methods}

ML methods have significantly enhanced the foundational framework of reactor physics. Moreover, theoretical advancements have addressed various challenges inherent in traditional nuclear reactor domains. Progress in steady-state problem resolution is expected to accelerate the application of ML in complex scenarios, including reactor operation simulations and safety design. Similarly, advancements in transient problem-solving will significantly improve the monitoring and maintenance of nuclear power plants.

The burnup issue, a domain intricately linked to both steady-state and transient phenomena, represents a critical area of study. Conventional analyses often prioritize the prediction of specific phenomena or the management of isolated variables, frequently overlooking the aspect of interpretability. However, accurate prediction of burnup can substantially improve the interpretability of these analyses. Research in this area holds the potential to optimize fuel management, support dynamic decision-making, and advance the study of nuclear fuel cycles.

Additionally, this work elucidates several ML-driven solutions to broader challenges in the field, such as issues of interpretability and advancements in radiation transport equations. These innovations are poised to significantly influence the future development of reactor physics.

\subsection{Steady-State Problems}

The steady-state problem is significant in nuclear reactor physics, focusing on the parameters or distributions of key variables as the system approaches its final reaction state. In the context of solving the governing equations in reactor physics, ML technologies play a crucial role. In steady-state problems, neutron diffusion often involves considerations of multiple energy levels and heterogeneous media, thus leading to heterogeneous neutron diffusion issues. Additionally, within the governing equations of reactor physics, predicting the k-eigenvalue of the reactor can be achieved without explicit solving, known as the k-eigenvalue problems.

Reactor performance refers to various factors such as neutron flux distribution, fuel utilization efficiency, and thermal-hydraulic behavior. Key parameters influencing reactor performance include temperature distribution, flow rates, and pressure conditions within the reactor core. The reactor performance prediction model integrates a reactor physics model and a thermo-hydraulic model, which allows for the prediction of thermal-hydraulic parameters such as temperature distributions in the core.

\subsubsection{k-eigenvalue Problems}

The k-eigenvalue problems based on diffusion theory have several applications of ML techniques. One of the earliest applications of ML in predicting the neutron diffusion equation dates back to 1999. In this research, a query-based adaptive retraining approach was developed, which involved reconstructing a two hidden-layer neural network for the rapid prediction of the eigenvalues of the fundamental mode of the neutron diffusion equation, specifically the effective multiplication factor ($k_{\text{eff}}$) \cite{bib:95}.

Based on their earlier work \cite{bib:75}, Elhareef et al. \cite{bib:78} investigated solving the k-eigenvalue diffusion equations in a multiregional configuration subject to a set of Robin boundary conditions, and the relative percentage errors of the two example k-eigenvalues are only about 0.77\% and about 1.2\%. They further investigated the forward PINN method, applied to the solution of the neutron diffusion equation for single and multi-energy groups, by using a freely learnable parameter to approximate the eigenvalues and a new regularization technique to exclude zero solutions from the PINN frameworks \cite{bib:43}. The computed results are compared with FEM solutions of the neutron fluxes and power iteration solutions of $k_{eff}$. The errors in the fluxes range from 0.63\% for a simple fixed-source problem to ~15\% for a two-set k-eigenvalue problem. The predicted k eigenvalue deviations from the power iteration solver range from 0.13\% to 0.92\%. To solve the problem of solution infinity when calculating the neutron diffusion equation by deterministic methods, Li et al. \cite{bib:96} took the power calculation equation as a supplementary equation to form a system of differential-integral equations with the neutron diffusion equation, and designed a coupled numerical and machine learning (ML) algorithm based on the finite-volume method of gradient updating for the system of equations, which is applied to the calculation of the neutron flux densities and volumetric heat rates of the energy groups at specified powers.

In a recent study, Yang et al. \cite{bib:7} proposed a data-enabled PINN (DEPINN) to solve the neutron diffusion eigenvalue problem. They \cite{bib:76} also performed uncertainty analysis and further numerical studies, and the proposed interval loss function to deal with noisy a priori data items greatly improve the robustness of DEPINN to noise. The method is expected to solve practical nuclear engineering problems in the field of nuclear reactor physics in the presence of observed data (with noise). Furthermore, they also studied Generalized Inverse Power Method Neural Network (GIPMNN), Physics-Constrained GIPMNN (PC-GIPMNN), and Deep Ritz Method (DRM) to solve the k-eigenvalue problem in neutron diffusion theory, among which the PC-GIPMNN method performs the best, and the PC-GIPMNN method has the best performance and the accuracy of ML method is very good in all the cases, the ML method accuracy outperforms FEM and is not affected by meshing \cite{bib:48}.

\subsubsection{Reactor Performance Prediction}

As described in the introduction, small reactors are currently being developed as the fourth generation of advanced reactors \cite{bib:4} focusing on reactors such as TFHR. These small reactors are usually located in remote areas for energy supply \cite{bib:27}. To achieve autonomous control of reactor systems with minimal intervention—thereby enabling long-term, stable, and efficient operation of nuclear reactors while reducing costs—Zeng et al. \cite{bib:27} investigated the autonomous control of small reactors. Their study focused on predicting the system's performance in a steady state. They developed a ML-based performance prediction model that integrates a reactor physics model and a thermo-hydraulic model, trained using the SVR method. This model was designed to predict reactor core behavior and its future evolution. The system comprised two main components: a prediction module and a decision-making module, which worked together to accurately and efficiently assess the reactor's future operational conditions.

\subsubsection{Temperature Field Prediction}

The prediction of temperature also serves as a key to the optimization of nuclear reactors, and with the help of accurate prediction, we can achieve temperature regulation. In the field of nuclear power generation, reactors with heavy water coolant and moderator are widely used, and the monitoring of the moderator state is crucial to ensure the proper operation of the plant \cite{bib:26}. Starkov et al. \cite{bib:26} focused on the prediction of the temperature field of the nuclear reactor moderator under steady-state conditions and established a prediction system for the temperature of the heavy water moderator. They predicted the temperature distribution in the core cross-section based on a cellular neural network architecture. Based on this, continuous optimization of the moderator can be achieved.

Additionally, related studies have focused on predicting critical conditions, particularly temperature. Park et al. \cite{bib:54} employed a ML model to predict wall temperature at critical heat flux (CHF), a crucial parameter in nuclear reactor heat transport. CHF marks the transition from nucleate boiling to transition boiling, serving as a key indicator of the reactor's performance under steady-state operating conditions in terms of heat transport efficiency.

\subsubsection{Power Prediction}

In the operation of a nuclear power plant, ensuring a homogeneous radial power distribution in the reactor core is essential to avoid overheating locally and to ensure safe operation. The study \cite{bib:97} used the group data processing method (GMDH) to predict the axial power and axial offset of the reactor core, as well as these parameters for the fuel assembly. By using the GMDH algorithm, the researchers were able to reconstruct the axial power offset across the reactor core from the in-heap detector signals and determine the optimal relationship between the detector signals and the core axial power and axial offset \cite{bib:97}.

The reactor's performance can also be effectively assessed by predicting the power distribution at a steady state. This is crucial because the power distribution directly influences the temperature distribution, as well as the transport and migration of fission products within the reactor. In \cite{bib:98}, lattice physics calculations was used to generate data to find analytical expressions for the radial power distribution in fuel cores, with the help of ML techniques to find representative expressions. Daniell et al. \cite{bib:99} designed and compared several Multi-Stage DNN (MSDNN) architectures for predicting the final steady state power of a research nuclear reactor after power variations. The results show that MSDNN models tend to perform better than standard feed-forward neural networks in terms of accuracy and generalization.

The use of DNN continues to be a viable alternative to standard physics modeling in nuclear systems and engineering. Alam et al. \cite{bib:32} attempted to develop a multi-stage prediction model consisting of 2 feed-forward DNNs to determine the final steady-state power at each transient in a real reactor facility. They separately output a resolution representing the large frequency band power output, which in turn is used as an input to realize the prediction. Physical information is collected from the initial conditions of the transient as well as the final steady-state \cite{bib:32}. The final combined model given achieves a classification accuracy of 96\% and an absolute prediction accuracy of 92\%.

Furthermore, Wan et al. \cite{bib:110f} proposed a neural network–based variable universe fuzzy control method for regulating power and axial power distribution in large PWRs. By establishing fuzzy inference rules based on a two-node reactor dynamics model and adopting a variable universe approach to design the input scaling factor—optimized online via a neural network—the method achieves precise control of reactor power and spatial distribution. Simulation results indicate that, under transient conditions, the control performance of this method is significantly superior to that of conventional fuzzy controllers.

\subsubsection{Power Peak Factor Estimation}

The problem of computing the PPF is directly related to the thermal stresses and potential local risks of nuclear fuel rods. In steady-state conditions, it reflects the ratio of the highest local power density to the average power of the core reached by the nuclear reactor after a long period of operation. To address the problem of accurate estimation of the PPF, Bae et al. \cite{bib:21} developed a model to process the signal and predict it accurately using measured signals from the reactor cooling system. The model is based on the SVR method and determines the NN weights by solving a quadratic programming problem.

In practice, this approach demonstrates superior generalization ability compared to traditional ANN when addressing nonlinear problems, resulting in lower prediction errors. In the context of solving the fuel loading problem, both Trontl et al. \cite{bib:22} and Hedayat et al. \cite{bib:100} focused on optimizing the PPF, which in turn facilitated precise regulation of core fuel loading. Tayefi et al. \cite{bib:23} further optimized fuel management strategies to enhance operational efficiency and safety, ensuring that the nuclear reactor operates within optimal conditions over long-term use. Additionally, they utilized the Hopfield neural network (HNNA) to identify the optimal distribution of axial enrichment variations, which effectively reduces peak neutron flux and power peaking factors \cite{bib:101}.

\subsubsection{Core Loading Pattern Optimization}

The prediction of a core loading pattern, or the reload strategy, is a typical reactor core design task that involves optimizing the material composition, geometrical arrangement, and thermal-hydraulic parameters of the core. This optimization process must also account for constraints related to economic considerations, safety standards, and nuclear safeguards. Such strategies aim to reduce fuel cycle costs while ensuring the reactor operates safely and efficiently. Conventional optimization usually consists of nuclear physics computational codes that rely on engineering judgments, heuristic rules, optimization algorithms, and evaluations, which are typically computationally intensive. To ameliorate this problem, Trontl et al. \cite{bib:22} evaluated the core loading pattern quickly with the help of SVR methods, thus significantly reducing the computational time.

Hedayat et al. \cite{bib:100} explored the idea of using ANN to develop a rapid evaluation system for assessing and optimizing core parameters during reactor fuel reloading, which are essential to ensure optimal performance and compliance with safety standards during long-term operation.

\subsection{Transient State Problems}

Transient simulations, particularly those dealing with reactor core dynamics during operational changes or disturbances, require a detailed understanding of reactor kinetics, thermodynamics, and control mechanisms. Advanced models are crucial for predicting and mitigating the effects of transient events that can impact reactor stability and safety.

Precise prediction of reactor neutron dynamics not only offers a more reliable theoretical foundation for core nuclear design, but also serves as a critical reference for guiding the safe operation of actual reactors, thereby helping to prevent core transient accidents.

A transient is defined as an event during which a nuclear plant transitions from one state to another state, the latter may be abnormal. Based on diffusion theory, Fick's law serves as a P1 approximation of the neutron transport equation, commonly referred to as the telegraph equation \cite{bib:61a}. However, this formulation neglects the time derivative term of neutron flow density, limiting its ability to accurately depict the actual neutron transport process during transient events \cite{bib:62}. Theoretical models for transient calculations primarily include point reactor kinetic (PRK) models based on diffusion equations, as well as nodal block diffusion neutron kinetic calculation methods that incorporate component homogenization \cite{bib:102}.

Key factors influencing transient behavior include Doppler reactivity, which responds to changes in fuel and coolant temperature. Traditional models, such as PRK models, are commonly used for transient calculations, but they often have limitations in representing the full dynamics of the reactor during transient events. To address these, more advanced models, including those based on nodal block diffusion neutron kinetics, are employed for more accurate predictions. Furthermore, the integration of machine learning (ML) models has shown promise in enhancing the accuracy of transient prediction and reactor control, allowing for better decision-making during these critical periods. These advanced methods can provide more reliable simulations of transient reactor behavior, which ultimately supports safer reactor operation and management.

\subsubsection{Doppler Reactivity}

The Doppler reactivity coefficient is one of the key safety parameters of the core, determined during the design phase alongside the moderator reactivity coefficients. In 3D reactor core behavior analysis, these reactivity coefficients are predicted but are not required by the solvers. The solvers consider local neutronic and thermal-hydraulic feedback through nodal or homogenized cross-sections provided to them. Only point kinetics models (PKM) require explicit input of reactivity coefficients. However, PKM is inherently fast, and therefore, there appears to be no need to replace it with an AI-based model to enhance its speed. Choi et al. \cite{bib:103} developed a comprehensive reactor dynamics model incorporating point kinetics equations (PKE), a thermodynamics model for hot water (TH), and a xenon (Xe) dynamics model to facilitate the automatic control of small modular reactors (SMRs). In a related study, Dorde et al. \cite{bib:33} proposed a method for calculating Doppler reactivity in conjunction with the PRK model for the Sodium Fast Reactors (SFRs), which was based on the ANN and yields similar or even better performance.

Furthermore, monitoring the startup transients of the Purdue University Reactor-1 (PUR-1) was conducted using a reactivity feedback scheme. This involved the use of experimental parameters derived from a neutron source, providing valuable insights into the reactor's behavior during initial startup conditions. The integration of advanced modeling techniques and real-time monitoring is essential for enhancing the safety and operational efficiency of modern nuclear reactors. These studies underscore the importance of accurately calculating Doppler reactivity and the potential benefits of incorporating machine learning methodologies in reactor dynamics.

\subsubsection{Behavior Prediction}

Reactors undergoing rapid changes or transient events need to have their behavior predicted. Traditional numerical methods are time-consuming and susceptible to accumulated rounding errors or floating-point overflow in the transient state itself. Therefore, Pirouzmand et al. \cite{bib:71} proposed a cellular neural network-based simulation of complex nuclear reactor behavior. Specifically, the method simulates the time-space response of the flux distribution in a two-dimensional Cartesian geometry, which in turn solves the problem of behavior prediction. The model is well stabilized and can be accurately analyzed to provide fast and accurate solutions for complex and rigid models.

Park et al. \cite{bib:104} proposed an evolutionary algorithm, the GMDH algorithm, to process and analyze real-time data obtained from detectors inside the reactor to reconstruct and monitor the axial power distribution of a nuclear reactor. The algorithm is robust to different power distributions and can cover the entire reaction cycle, thus monitoring and evaluating the behavior of the nuclear reactor in real-time. 

Utilizing extensive data from the Multi-Application Small LWR (MASLWR) integrated test facility at Oregon State University, Gomez Fernandez et al. \cite{bib:25} developed a NN topology designed to simulate and predict the behavior of a nuclear reactor during a loss-of-feedwater transient (LOFW). This NN model effectively captures the dynamic changes in various physical parameters within the reactor during the LOFW event, providing a robust tool for analyzing reactor response under such critical conditions.

In the previous subsection on the application of ML methods in steady state, Zeng et al. \cite{bib:27} investigated the prediction of steady-state properties for small reactors, and similarly, for the transient problem of micro reactors, Mendoza et al. \cite{bib:105} proposed a power transient module that reliably identifies reactor operation as steady state, power rise, or power fall using data from a large-scale simulator. They also explored various data-driven methods to accurately assess power transients, namely principal component analysis, SVM, DNN, and CNN \cite{bib:105}.

To identify the state of the nuclear reactor, Zubair et al. \cite{bib:106} developed a model that includes the primary and auxiliary systems of the nuclear power plant. In addition, it includes models of the control system and instrumentation used to monitor and regulate the reactor as an important component of data extraction and transient modeling as an integral component of data extraction and transient modeling. Based on the data generated by Western Services Cooperation's General purpose PWR (GPWR) simulator as well as MATLAB tools, they \cite{bib:106} simulated a total of nine different transient events under twelve different initial conditions to create a dataset with 72,000 observations. Nine types of classification models (a total of 33 predefined models) were trained and validated using a classification learner application. The Neural Network Classifier had the highest average accuracy of 90\% in the validation of the ML-based approach. The study maintained a low execution and computation time while maintaining high accuracy. 

In a further study, to optimize the model, the study \cite{bib:107} integrated feature selection using the minimum redundancy maximum relevance (mRMR) algorithm. As a result, the KNN model achieved 97\% of the highest accuracy in the primary system, while the effective linear discriminant and logistic regression models obtained 99\% of the highest average accuracy in the secondary system \cite{bib:107}.

\subsubsection{Forward and Inverse Problems}

Real-time estimation of PPF is also important under transient conditions. This is needed to capture the rapid changes in PPF due to changes in operational variables, which include control rod movement, etc., to accurately detect and respond to special safety issues. Bae et al. \cite{bib:21} investigated the capture of transient PPF changes based on the SVR method. Focusing on the VVER-1000 reactor core power distribution problem, Pirouzmand et al. \cite{bib:24} estimated the relative power distribution and PPF of the reactor core with the help of ANN. They first monitored the radial RPD of the core and screened the axial relative power of the fuel assembly to detect the PPF of the core.

To achieve real-time power prediction, Gong et al. \cite{bib:38} proposed a reactor physics digital twin model that simultaneously solves the positive and negative problems of predicting power-given parameters and additional power field measurements to solve the input parameters of the operation. They \cite{bib:38} developed a non-intrusive forward model utilizing a Singular Value Decomposition (SVD) Autoencoder (AE) downscaling model, in conjunction with machine learning methods such as KNN and decision tree. The test results demonstrated that this approach outperforms traditional reduced-order methods, such as Proper Orthogonal Decomposition (POD). Regarding the inverse problem, they investigated the generalized latent data assimilation method to achieve real-time construction of the inverse model characteristics by evaluating the forward model multiple times \cite{bib:38}. Furthermore, Gong et al. \cite{bib:10} reconstructed the reactor's high-fidelity neutron field rapidly with real-time input parameters based on the online situation. They simulated different scenarios such as burnup, control rod insertion step, power level, and coolant temperature over the life cycle of the HPR1000, demonstrating the role played by transient problem-solving in online monitoring.

\subsubsection{Response Decision}

Anil et al. \cite{bib:108} predicted future reactor states under possible control strategies especially for reactor systems under transient conditions to evaluate and determine the optimal control strategy for autonomous control of nuclear reactors. The prediction model, based on initial and boundary conditions, can predict the fuel centerline temperature with predictive uncertainty and predict the future reactor state for all possible control strategies to select the optimal control strategy in the face of contingencies.

Konstantinos et al. \cite{bib:77} used a point kinetic model as a basis for predicting the neutron density in transients with the help of transfer learning PINN (TL-PINN). This is an alternative to traditional reactor transient modeling. The transfer learning achieves a significant performance improvement compared to PINN, greatly reducing the number of iterations for model training. They generated transients with different ranges of experimentally relevant variables using PUR-1 experimental data.

Since the extreme stress of operators in most nuclear accident scenarios is prone to operational errors, ML models can provide real-time support and advice and act as expert systems to assist in accident handling and decision-making. Based on this idea, Nguyen et al. \cite{bib:109} developed a multivariate time series ML meta-model to predict the transient response of a nuclear power plant experiencing steam generator tube rupture (SGTR). The model employs a hybrid CNN-LSTM model in recurrent neural networks (RNNs). They also implemented Bayesian Neural Networks (BNNs) to address the uncertainty in the prediction. The final model successfully captured the latent features of the data, and the BNN-LSTM approach provided additional monitoring of the level of uncertainty associated with the heap prediction \cite{bib:109}.

The successful application of LSTM extends beyond this point. Bae et al. \cite{bib:110} proposed a data-driven prediction model which is based on a multi-step prediction strategy and ANN, by testing MLP, RNN, and LSTM respectively, they finalized a multi-input multiple-output strategy and LSTM, which successfully dealt with multivariate problems under multiple emergencies of incident handling. The model not only achieves prediction accuracy, but also a guarantee of accident transient prediction capability, with prediction time consumed averaging only 0.06 seconds.

\subsubsection{Thermo-hydraulic coupling}

The nuclear reactor is a complex system with a wide range of physical fields such as neutron density field, flow field, temperature field, etc. widely present in the reactor system \cite{bib:110a}. There are interactions between these physical fields, so there is a strong coupling relationship between them. And these coupling relations are not negligible during reactor operation, especially during transient processes. Additionally, for neutronics, the neutron fluence rate determines the distribution of fission power inside the core, which affects the temperature field of the fuel rods in addition to the one of water, and ultimately changes the whole temperature field. As for thermal hydraulics, the thermodynamic parameters strongly affect the macroscopic cross section. The interaction of these physical fields forms the basis for the coupling of neutronics and thermo-hydraulics \cite{bib:110b}. Therefore, we will broadly introduce here the machine learning applications in the scenario of reactor thermal hydrodynamics coupled with reactor physics.

In recent years, the application of ML techniques in nuclear reactor thermo-hydrodynamic coupling and related areas has achieved remarkable progress, offering innovative solutions for the real‐time prediction, optimization, and control of complex systems. For instance, Zhang et al. \cite{bib:110c} proposed a neural network–based rapid prediction method for multi-physics field coupling in heat pipe reactors. By replacing traditional numerical modules with neural networks, this method achieves high-precision coupling predictions for multiple physical domains (e.g., neutronics, thermal, and mechanical) while reducing computation time from several hours to less than four minutes—without compromising key parameter accuracy (e.g., maximal stress difference <2 MPa, average fuel temperature difference <3 K). Moreover, the approach requires relatively modest training data and permits dynamic model correction and optimization based on prediction performance, thereby offering significant support for accident early warning and on-site deployment. 

Traditional approaches often decouple these phenomena, for instance by using one-way coupling where neutronics and thermal-hydraulics provide boundary conditions for fuel performance analysis. However, this simplification may overlook essential core-wide interactions. To address this, Che et al. \cite{bib:110x} developed a machine learning–assisted surrogate model that fully integrates coupled neutronics, thermal-hydraulics, and fuel thermo-mechanics. By combining look-up tables with advanced ML algorithms—including rule-based techniques for effective feature extraction—the surrogate model accelerates simulation time by up to four orders of magnitude relative to conventional codes (e.g., FRAPCON) while preserving key accuracy metrics. This approach enables more realistic, full-core predictions and facilitates tighter integration of fuel performance into core design optimization.

In addition, Aghili Nasr et al. \cite{bib:110y} tackled a core challenge in coupled simulations: dynamically updating neutron cross sections as a function of evolving state variables. Their method employs Gaussian Process Regression (GPR) to substitute traditional logarithmic correlations for temperature-dependent cross section updates. Implemented within a multiphysics framework based on OpenFOAM, the GPR approach was validated on benchmark problems and a 2D axisymmetric Molten Salt Reactor configuration. The results demonstrate that, particularly for isotopes with prominent absorption resonances (e.g., $\mathrm{^{232}Th}$ and $\mathrm{^{238}U}$), the method significantly influences the effective multiplication factor ($k_{eff}$) and transient peak power predictions—highlighting the importance of robust coupling for reactor safety analysis.

Overall, these studies underscore the transformative potential of advanced surrogate modeling and ML techniques in capturing the nuanced interplay between reactor physics and thermal-hydraulics. Not only do these approaches substantially reduce computational burdens and enhance predictive fidelity, but they also offer flexible and efficient strategies for regulating reactor operation. The vast application potential of ML in improving multiphysics field coupling and thermo-hydrodynamic integration is pivotal for achieving safe, economical, and intelligent operation of nuclear power plants. For a comprehensive review of ML applications in thermal hydraulics, readers are encouraged to consult \cite{bib:111a}.

\subsection{Fuel Burnup Problems}
Traditional methods based on the burnup equation include higher-order Chebyshev rational approximation \cite{bib:111}, Krylov subspace methods \cite{bib:112}, etc. The direct use of ML technology in burnup calculation, burnup distribution prediction, nuclide density prediction, etc., has already achieved a large breakthrough.

Burnup calculations are critical to understanding and predicting changes in calculating nuclei inventories and mass flow evolution in an entire fuel cycle, from the mine to the final disposal \cite{bib:93}. Burnup modeling allows the simulation of the operating conditions of reactor units and the dynamics of fuel cycle facilities to optimize fuel management and improve the efficiency of nuclear energy use. Accurate burnup predictions are also important to ensure the safe operation of nuclear reactors and to support fuel cycle decision-making. In dynamic fuel cycle simulation tools, these isotopic compositions are unknown until the code is executed. Therefore, models are needed to calculate fuel burnup \cite{bib:93}. The burnup prediction in turn includes the burnup distribution \cite{bib:113} as well as the nuclide density distribution \cite{bib:114}. In addition, ML techniques have some applications to other problems \cite{bib:35}.

\subsubsection{Optimization of Burnup Calculation}

To simplify and speed up the process of pressurized water reactor burnup calculation, Leniau et al. \cite{bib:93} proposed a method to generate a pre-calculated set of fuel burnup calculations. They developed a NN-based approach. Their method \cite{bib:93} was applied to the dynamic fuel cycle simulation tool CLASS to predict the Pu content and the average cross-section in the fuel through a neural network, which significantly improved the computational efficiency and accuracy. The method was able to complete burnup calculations in less than a minute and had an average error of only 0.37\% in predicting Pu content, demonstrating its potential for nuclear fuel cycle management and burnup prediction. 

On this basis, Courtin et al. \cite{bib:115} investigated the MOX extended fuel nuclide prediction. Their proposed fuel loading model based on infinite multiplication factor ($k_{\text{inf}}$) calculations predicts the amount of Pu in the fuel required to reach the target burnup, as well as the enrichment of U in specific cases, with an accuracy of 270 pcm and a deviation of less than 4\% for the primary nucleus at the end of the cycle (EOC).

To perform safeguards, the spent nuclear fuel needs to be checked for the correctness of the fuel assembly declarations and the completeness of the relevant declarations for each fuel assembly, i.e., it is determined that the spent fuel assembly does contain nuclear material and that no diversion of a part of the assembly has taken place \cite{bib:116}. Based on simulated pressurized water reactor fuel data, Grape et al. \cite{bib:116} used random forest regression to explore the predictive power of fuel parameters for initial enrichment, burnup, and cooling time.

\subsubsection{Burnup Distribution Prediction}

Accurate prediction of burnup distribution is also crucial for fuel reloading design and optimization, which is essential for the safety and economy of the reactor. However, due to the heterogeneous flow distribution in the reactor core and the manufacturing deviation of each component, there is a deviation between the actual and the simulation, leading to an error in the power distribution that is subsequently transferred to the burnup distribution \cite{bib:113}. Guo et al. \cite{bib:113} proposed a data assimilation method. They first combined the measured values of power distribution with the results of numerical simulation, established the relationship between burnup distribution and power distribution with the help of ANN, and finally calibrated the burnup distribution with the help of the 3D variational method. The method was validated on the CNP1000 PWR, and the error of the burnup distribution was significantly reduced.

\subsubsection{Nuclide Density Prediction}

A related study developed and validated a DL-based model for predicting burnup nuclide densities. To predict the nuclide densities of ${}^{235}\text{U}$, ${}^{239}\text{Pu}$, ${}^{241}\text{Pu}$, ${}^{137}\text{Cs}$, and ${}^{154}\text{Nd}$, Lei et al. \cite{bib:114} used the mean square error (MSE) of the DNN as a loss function by comparing it with the relative error of a multilayer perceptron model. They found that DNN overcame both the problem of excessive prediction error of traditional ML algorithms in the low burnup region and performed better in the middle and high burnup regions, proving the feasibility of ML techniques in nuclide density prediction.

\subsubsection{Other Fields in Burnup}

In order to analyze the equilibrium burnup state of molten salt reactors (MSRs), Chen et al. \cite{bib:35} generated equilibrium and non-equilibrium state neutron information of fuel assemblies with different geometrical parameters as a dataset, and quickly filtered the models for the equilibrium burnup state and predicted the neutron fluxes and multiplication ratios with the help of 15 different ML regression algorithms. In the results, the LGBM model performed the best.

Soto et al. \cite{bib:117}found that with the help of ML methods, it is possible to achieve a significant improvement in accuracy over a short fuel cooling time, increased robustness to spectral compression, and competitive burnup prediction even when only background signals are used, which is difficult to achieve with conventional methods.

Burnup of nuclear reactors on the other hand also by the interaction between the fuel and the cladding, which is currently the subject of more frontier studies \cite{bib:118}. The literature indicates that the ML method is expected to be used in conjunction with energy dispersive x-ray spectroscopy (EDS) analysis of scanning electron microscopy collected in this study for the analysis of high-magnification back-scattered electron (BSE) datasets to facilitate the rapid and consistent extraction of statistical information on fuel-cladding chemical interaction (FCCI).

\subsection{Other Problems}

\subsubsection{Observable Measurement Descriptions}

Accurate definitions of nuclear physics observables become difficult since the information from differential nuclear physics experiments and theories is usually subject to large uncertainties, thus making the precise definition of nuclear physics observables challenging. On the other hand, the overall experimental data representing the applications of these observables are usually more precise. At the same time, there is too much experimental data to identify problems in the observables through human expert analysis alone \cite{bib:55}.

For the description of these quantities, Neudecker et al. \cite{bib:55} proposed an analytical approach to identify groups of observable measurements in nuclear physics that may lead to biases in integral experimental simulations using the Random Forest algorithm and the SHAP metric.

In a practical application, the study characterizes the fission observable measurements of ${}^{241}\text{Pu}$. They \cite{bib:55} combined differential experimental data and theoretical modeling to determine that the ${}^{241}\text{Pu}$ fission cross-section may be the main cause of simulation bias. The high-energy interval data for ${}^{239}\text{Pu}$, which could not be accurately described, was still able to narrow down the range of nuclear data that could be problematic. This demonstrates the successful application of ML technology in the description of observed quantities and its wide prospects.

\subsubsection{Radiation Transport Problems}

The radiation transport problem covers various types of radiation, including neutron transport in reactor physics. It describes the flux of radiation through a heterogeneous medium. For the radiative transport problem, traditional ML methods are difficult to apply due to the large computational cost of creating data through simulation. Huhn et al. \cite{bib:53} investigated the PINN method to solve the radiative transport problem in the absence of data. They also proposed the application of Fourier features in the algorithm, which yielded good performance on heterogeneous problems. For sampling, they investigated heuristic-based methods. In a practical set of one-dimensional radiative transport problems, PINN produces consistency with the traditional method solutions under fine grids \cite{bib:53}.

\section{Reactor Physics: Data and Applications}

The previous sections have offered a thorough exploration of how ML technology can enhance various facets of reactor physics, encompassing the architecture of governing equations, state parameters, and their application in both steady-state and transient problems. This section will shift the focus to the impact of ML methods on performance improvements in real-world applications. Specifically, it will examine advancements in data processing and identification—referred to as "data in modeling"—as well as their roles in simulation, online monitoring, and design within industrial contexts.

ML models work by learning patterns from large datasets. In reactor physics, these models are trained on historical data from reactor simulations and real reactor data. After training, the models can predict reactor behavior, optimize operational parameters, and even detect anomalies in real time. These models typically rely on supervised learning techniques, where the model learns from labeled data (e.g., reactor states and corresponding outcomes) and generalizes to make predictions for unseen scenarios.

Despite the potential of AI in reactor physics, there are several challenges. First, the complexity of reactor dynamics, which involve numerous interacting physical phenomena, makes it difficult for models to generalize well across different reactor types or operational conditions. Additionally, training ML models requires high-quality, well-labeled data, which can be scarce or difficult to obtain in real-time. Moreover, many ML models function as black boxes, making it hard to interpret how they arrive at specific decisions, which can be a significant concern in safety-critical applications like nuclear reactors.

While ML models have shown promise in operational simulation and reactor monitoring, their accuracy is often limited by the fidelity of the data they are trained on. In many cases, ML models are used in conjunction with traditional physics-based simulations to ensure the safety and robustness of predictions. For example, hybrid models combine machine learning with high-fidelity reactor physics simulations to reduce errors associated with low-fidelity models. Moreover, real-time retraining of ML models is sometimes necessary to account for changes in reactor conditions or to update the models with new data, further complicating their deployment.

\subsection{Data in Modeling}

ML methods also offer significant benefits for data processing and enhancement tasks, such as model data correction, data generation, data validation analysis, and noise diagnosis. These aspects are not directly related to steady-state or transient problems but involve optimization during the model preparation stage. As such, they are discussed in the section addressing additional challenges.

\subsubsection{Model Correction Problem}

While high-fidelity transport calculations are certainly accurate, their large number of calculations makes them difficult to apply, especially in the implementation of forecasting, where it is almost impossible to get direct access to high-fidelity model/data in real-time. Therefore, the study moves towards the consideration of error correction for low-fidelity model/data.

For operational simulation and real-time prediction in boiling water reactor (BWR), Oktavian et al. \cite{bib:119} investigated machine learning (ML) methods. Specifically, they generated initial results using a standard two-step simulation \cite{bib:110} process and then used a ML model to correct for errors between high and low-fidelity data. To correct the errors between the nuclear reactor mechanical model and the actual reactor, Wang et al. \cite{bib:120} proposed a hybrid mechanism and NN approach to nuclear reactor modeling to eliminate or minimize these deviations. Specifically, based on simulation data from RELAP5, they built a point reactor dynamics model as a model of the core and calibrated it using genetic algorithm (GA) to optimize the BPNN reactor reactivity feedback.

\subsubsection{Nuclear Data Validation Analysis}

Nuclear data validation analysis focuses on identifying possible problems in nuclear data through ML models. This process focuses on the overall quality and predictive power of nuclear data. As mentioned in literature \cite{bib:56}, ML methods assess the reliability of a large number of nuclear data features and corresponding experimental results by analyzing them in the simulation of critical nuclear experiments. These models are based on measurements and simulations of selected key components from the International Criticality Safety Benchmark Evaluation Project (ICSBEP). The $k_\text{eff}$, along with its sensitivity to nuclear data and baseline features, are used as inputs to the Random Forest (RF) regression model. The RF model is employed to encode the complex interdependencies between factors such as sensitivity profiles, descriptive measurements of the nuclear data, and both simulated and experimental $k_\text{eff}$ values. This approach aims to identify which features of the nuclear data are most influential in predicting deviations, thereby enhancing our understanding of the underlying relationships.

A related study \cite{bib:86} also focuses on the potential role of ML in supporting cross-section evaluation and proposes a data enhancement model for neutron-induced response cross-sections. Specifically, they fit nuclear data from the EXFOR database based on decision trees and KNN to predict data predictions for important nuclides, and the final predictions match the measurements very accurately.

\subsubsection{Data Synthesis Problem}

Before building a nuclear power plant, the proliferation potential of a nuclear reactor needs to be understood. The digital twin is the solution to identify the behavior of nuclear proliferation reactors. Based on the parameters of the AGN-201 reactor of Idaho State University, Palmer et al. \cite{bib:121} discussed data synthesis techniques to generate data to train the digital twin model. Additionally, they trained LSTM on synthetic data to further validate the application of these methods in the implementation of digital twins. The benefit of this approach is that it effectively captures the aging and sensor drift of equipment caused by environmental factors and other influences. By monitoring the error between the actual data and the synthesized data, it can signal the need for reprocessing the synthesized data at the appropriate time, ensuring more accurate and reliable monitoring and analysis over the equipment's lifespan \cite{bib:121}.

\subsubsection{Noise Diagnostic Problems}

A nuclear power plant is a large and complex system equipped with numerous monitoring sensors. Within the reactor core, it is critical to detect anomalies as early as possible to prevent irreversible consequences. Although the core can only accommodate a limited number of sensors for data acquisition, neutron detectors remain highly effective in capturing perturbations, even those occurring at significant distances from the sensors. This capability is due to the intrinsic nature of fission and scattering reactions, as well as the transport of particles within the core, which allows for the detection of changes throughout the reactor system \cite{bib:122}.

Based on the capture of anomalous states, Kollias et al. \cite{bib:122} used ML method for noise capture and thus anomaly localization. Specifically, they performed semantic segmentation, classification, and localization for multiple simultaneous perturbations with the help of DL models. They further developed a domain adaptive approach and applied it to real reactor simulations.

\subsubsection{Data Fidelity Mapping}

To reduce the cumulative error of long-term prediction while minimizing the computational costs, Zhou et al. \cite{bib:123} introduced a novel prediction model known as the multi-scale physics-constrained neural network (MSPCNN). This model utilizes a multi-fidelity convolutional autoencoder (CAE) to map data of varying fidelity into a uniformly shared potential space. This approach serves two key purposes. First, it enables low-fidelity data to complement high-fidelity data during the training phase. Second, it allows for the enforcement of physical constraints within the low-fidelity domain, rather than at the high-fidelity level. This strategy significantly reduces the costs associated with offline data acquisition and preprocessing, while simultaneously ensuring model accuracy. This advancement represents a notable improvement over traditional PINN. Additionally, the MSPCNN model exhibits robustness to noise, providing reliable predictions even in sub-optimal conditions.

\subsection{Applications}

In this section, we explore practical applications of ML methods in reactor physics within real nuclear power plants. We focus on a series of technological innovations and methodological improvements brought to nuclear power plants by the application of ML technology, especially the advancements brought to industry by the enhancements in the field of nuclear reactor physics. We will mainly focus on operation simulation, online supervision and maintenance, and core optimization design based on real nuclear power plants.

\subsubsection{Applications in Operational Simulation}

Neutron simulation of reaction processes is essential for the design and operation of nuclear reactors, serving as both a validation of safety and operational capability. Efficient simulations not only provide operators with real-time data and critical metrics to optimize power distribution and burnup and extend equipment life, but also hold the potential to significantly accelerate the design of next-generation advanced reactors \cite{bib:124}. A considerable body of literature focuses on the application of ML methods in reactor simulations, which will be systematically reviewed and discussed in this section.

\noindent\textit{Massively Parallel Simulation Acceleration}

Neutron simulations rely on physical models and numerical methods that often require large amounts of data. Current codes for neutron simulation often rely on tabulated data and simplified physical models, which reduces runtime at the expense of accuracy. While high throughput Graphics Processing Units (GPUs) and GPU-accelerated High-Performance Computing (HPC) platforms can reduce this dependence. Optimizing neutron codes and their algorithms is key to fully utilizing the advanced computational power and maximizing fidelity. Dorville et al. \cite{bib:124} presented three computational projects aimed at improving the performance of neutron transport codes through a modular approach. The common goal of these projects is to enable neutron codes to make optimal use of the massively parallel acceleration provided by GPUs.

\noindent\textit{Modeling and Simulation of Reactor Physical Phenomena}

Anil et al. \cite{bib:108} enhanced the efficiency of simulation predictions by employing ML techniques. These ML-driven methods integrate a physics-informed bootstrap mechanism, which effectively augments the model's capacity to learn the underlying physical phenomena through data-driven approaches. The approach constitutes a framework designed to guide and improve predictive modeling in ML-based autonomous control systems.

To enable simulation, Radaideh et al. \cite{bib:125} developed a novel modeling framework. The framework combines multi-physics simulations and real data, validated by an uncertainty quantification task and finally integrated by ML methods. The code can simulate different physical phenomena inside the reactor, such as neutron transport, reactor dynamics, fuel depletion, two-phase flow, and fuel performance \cite{bib:125}.

For the efficient conduct of the simulation process, Che et al. \cite{bib:126} also explored the application of ML methods in improving the computational efficiency of fuel performance analysis of nuclear reactor cores. They aimed to use ML techniques to accelerate full-core simulations by constructing fast-running surrogate models.

\noindent\textit{Fuel Reload for BWR}

Optimization of the fuel reloading problem for a BWR usually involves combinatorial optimization, which is an NP-complete problem. From expert systems to genetic algorithms, an accurate reactor simulator is required. However, conventional simulators require a significant amount of time, making optimization challenging to carry out efficiently and accurately \cite{bib:17}. 

It has already been proposed in the literature in 2003 to adapt, modify, or even replace the reactor simulator. Based on the emerging NN model, Ortiz et al. \cite{bib:17} proposed to replace the nuclear reactor simulator with this model to predict several variables in the BWR optimization of the fuel changeover process, including the $k_{\rm eff}$ of the reactor and safety-related thermal limits. After being trained, the NN can give results in a few seconds, which is certainly a huge improvement compared to the few minutes of the traditional simulation.

Based on previous work, Ortiz et al. \cite{bib:18} further investigated the optimization of the loading pattern in BWR, which was applied in finding the optimal reproduction pattern for five cycles at the Laguna Verde nuclear power plant in Mexico. Moreover, Trontl et al. \cite{bib:22} proposed the use of a support vector regression (SVR) model to quickly evaluate and optimize core loading patterns in nuclear reactors. This model, as a supervised learning method, can efficiently predict and evaluate different loading patterns and significantly reduce the computational time required by traditional optimization methods to improve the efficiency of core loading pattern evaluation in nuclear reactors \cite{bib:22}.

\noindent\textit{Operational Simulation of BWR}

During the operation of a BWR, the industry requires dynamic and precise control of the reactivity to maintain efficient operation. Throughout the reactor cycle, adjustments will be made to the insertion of control rods, and the flow rate of the core, and thus the reactivity will be kept under control \cite{bib:119}. Oktavian et al. \cite{bib:119} introduced a new methodology to use ML methods to enhance the simulation of BWR operation. The ML technique is primarily used to identify and correct problems in low-fidelity simulation results, aiming to produce a fit to high-fidelity data after correction of the data. The approach focuses on enhancing the regular BWR model without increasing the speed of the operation, which is critical for fuel management.

Oktavian et al. \cite{bib:70} further tested two different ML methods, including DNN and extreme gradient augmentation models, based on data from Monte Carlo simulations, validated in a 2*2 simulation of a simple BWR. They \cite{bib:127} also generalized the model to a small BWR with 88 fuel bundles and validated it in a full-core BWR with 560 fuel bundles based on Unit 1 of the Edwin Hatch nuclear power plant. They found that the model has good results for a range of conventional BWR operations with a fixed core design.

\noindent\textit{Fuel Loading of PWR}

The fuel is usually able to be modeled in a dynamic fuel cycle simulation tool \cite{bib:93} from mining through the entire fuel cycle until it becomes spent fuel at the end of the reaction, including the nuclear inventory present in it and the evolution of the mass flow. In this process, what needs to be studied is fuel consumption, which relies mainly on fuel-loading models and average cross-section predictors for processing \cite{bib:93}. For the case of PWR-MOX, Leniau et al. \cite{bib:93} investigated a neural network-based approach to build a loading model and a cross-section prediction model for loading MOX fuel. The former predicted the Pu content with an average error of 0.37\%, and the latter realized high-precision fuel calculation within one minute.

In addition, there are differences between numerical simulations and actual cores. These differences can lead to errors in the simulated values of the power distribution and burnup distribution. To reduce these errors, Guo et al. \cite{bib:113} proposed a data assimilation method for PWR burnup distributions, aiming to calibrate the burnup distributions using the power distribution measurements. Specifically, they used a three-dimensional variational (3DVAR) algorithm for burnup distribution calibration and applied an ANN algorithm to establish a complex relationship between the burnup distribution and the power distribution \cite{bib:113}. Through engineering validation, this data assimilation method was successfully applied to the CNP1000 PWR operating in China.

In the realm of pressurized water reactor (PWR) fuel loading optimization, Wan et al. \cite{bib:110d} developed a rapid evaluation model using an improved convolutional neural network integrated with the Inception-ResNet architecture. This model can accurately predict core parameters in approximately 0.0005 seconds and, when coupled with a genetic algorithm, efficiently identifies the optimal fuel loading pattern—completing the entire optimization process in about 20 minutes. This advancement not only enhances the operational safety of nuclear power plants but also improves their economic performance.

\noindent\textit{Optimization of HTR-ESS}

Based on the diffusion module in HTR-ESS, Li et al. \cite{bib:87} proposed a novel method, meeting the real-time requirements in the ESS. They further demonstrated through numerical experiments that the method is more accurate and efficient than existing methods.

On the other hand, due to the limited number of samples, the cross-section calculation model using the traditional model tree is easy to be overfitted in the leaf nodes, which leads to the wrong cross-section variation of the reactor core state parameters, which further appears to be inconsistent with the reactor physics laws. Moreover, the method fails at subspace boundary discontinuities. To overcome these two major difficulties, Tan et al. \cite{bib:91} improved the model with some novel methods and produced a more accurate cross-section as compared to existing methods and fulfills the requirement of real-time calculation. They also performed a validation of the suitability of the method with the help of a 10 MW high-temperature gas-cooled reactor (HTR-10) simulator.

\noindent\textit{Digital Twin in Simulation}

A digital twin is defined as a virtual representation of a physical asset, such as a reactor core, implemented through data and simulators to better understand the behavior of a nuclear reactor system. This concept is valuable for real-time prediction, optimization, monitoring, control, and enhanced decision-making \cite{bib:128}. In an early application in reactor physics, Gong et al. \cite{bib:38} proposed a reactor physics operational digital twin (RPODT) prototype for predicting neutron flux and power distribution in nuclear reactor cores for online monitoring purposes. They demonstrated the effectiveness of this digital twin using a real-world engineering case involving the HPR1000 reactor.

In early applications of reactors, one of the key components of RPODT is simulation modeling \cite{bib:38}. For modeling and simulation of reactor cores, the most unique and important physical fields are the neutron field and the associated core power distribution. Therefore, Gong et al. \cite{bib:38} further proposed a new digital twin model based on the ML approach, which solves both the forward problem and the inverse problem and was well validated in the core simulation of the HPR1000, and the accuracy is acceptable from an engineering point of view.

Furthermore, they \cite{bib:10} delved into digital twins by proposing a reduced-order model combined with ML methods. In the simulation of the reactor core modeled by two group neutron diffusion equations, the model parameters are influenced by input parameters over the life cycle of the HPR1000, they tested different burnups, control rod insertion steps, as well as power levels, coolant temperatures, etc., respectively. The study pointed out that ML-based predictive models need to be retrained each time for new reactor conditions, in which case real-time inference by digital twins (DT) would be prohibited \cite{bib:129}. To ameliorate this problem, Kobayashi et al. \cite{bib:129} investigated the feasibility of a deep neural operator network (DeepONet) as a robust agent modeling method in the context of DT-enabling technology for nuclear energy systems. The benefit of DeepONet is that it relaxes the requirement of continuous retraining, making it suitable for both online and real-time prediction components. Through testing and evaluation, the method does show an improvement over traditional methods in terms of both accuracy and efficiency. In addition, DeepONet has demonstrated generalizability and computational efficiency as an effective surrogate tool for DT components.

\subsubsection{Applications in Online Monitoring}

Safe operation and effective maintenance of nuclear reactors are critical. The development of economically beneficial and safe operations requires more accurate, comprehensive, and real-time analysis of neutron parameters \cite{bib:130}. A real-time monitoring system is needed to confirm that relevant safety requirements are not violated during reactor operation \cite{bib:131}. In the current literature related to online supervision, ML techniques are mainly applied in real-time monitoring, autonomous control, condition detection, parameter verification, signal monitoring, and fault monitoring and diagnosis of reactors. This section focuses on the ML approaches in these applications, and the practical benefits achieved.

\noindent\textit{Real-time Monitoring of Reactor Parameters}

The operation of nuclear reactors requires real-time monitoring and adjustment to ensure that safety limits are not exceeded under various operating conditions, which challenges the time complexity and accuracy of conventional methods to be harmonized. For the prediction of the PPF of a nuclear reactor, Bae et al. \cite{bib:21} proposed a model that is capable of predicting the PPF in a steady state and was also able to respond quickly to transient events. This in turn allows for a better understanding of the behavior of the reactor under different operations and thus more accurate decision-making during supervision by the operator. They also extended the application to the core operation limit supervisory system (COLSS) as well as the core protection calculator system (CPCS), further improving the accuracy and reliability of the system \cite{bib:21}.

Pirouzmand et al. \cite{bib:24} constructed a real-time monitoring system to predict the neutron parameters of the VVER-1000 core. Specifically, the training process includes obtaining different operating states by controlling different power density distributions of the control rods, and then training the MLP for each state. Based on the neural network established by the above training, the method also utilizes the signals from the out-of-core neutron detectors and the core parameter information for the prediction of axial and radial relative power distributions and the PPF.

In-core power distribution is one of the most important safety parameters in the operation of nuclear power plants \cite{bib:104}. The existing methods usually utilize detectors to measure a modest amount of the signal, and use a fixed basis function; in this way, the accuracy of the power distribution reconstruction depends on the estimation of the fitting parameters \cite{bib:104}. Park et al. \cite{bib:104} proposed the Group Method of Data Handling (GMDH) analysis to give the optimal form of the basis function of the nuclear power shape.

\noindent\textit{Monitoring of Moderator Condition}

Nuclear reactors generally have water coolants and moderators in them. Accurate monitoring of the moderator plays a crucial role in ensuring the proper operation of the power plant \cite{bib:26}. Based on the reactor model designed for the CANDU Darlington heavy water reactor as training data, Starkov et al. \cite{bib:26} proposed cellular NNs to consider the feasibility of real-time temperature prediction for the center section of the drain vessel. Their model can analyze events inside the drain vessel and assess the asymmetry of the heavy water volume heating, as well as evaluate the motion and mixing direction of the moderator flow. By using a committee of developed cellular structures, temperature anomalies can be estimated for the entire reactor volume, which can form the basis for full-scale 3D simulations of moderator conditions in the calandria \cite{bib:26}.

\noindent\textit{Nuclear Reactor Autonomous Control}

Autonomous control of nuclear reactors is also one of the supervisory applications of nuclear reactor physics. The most important aspect of autonomous control is prediction. In the field of nuclear engineering, prediction is the process of predicting the future condition of a system or equipment based on the current signs and symptoms of a failure \cite{bib:108}. An accurate prediction is important for strategy selection in failure situations. Anil et al. \cite{bib:108} built a framework to guide the development and evaluation of ML-based prediction models for autonomous control systems.

Additionally, Zeng et al. \cite{bib:27} developed a core behavior prediction model based on the Thorium-Fueled High-Temperature Reactor (TFHR), which integrates a thermal-hydraulic model and a reactor physics model. They utilized RELAP5-3D simulations, driven by DAKOTA, to simulate various operating conditions and accident scenarios, such as different reactivity insertion rates and insertion times, to generate the dataset. Experimental results demonstrate that this methodology effectively predicts and assesses reactor states, enhancing system response and efficiency, and thereby guiding the autonomous control of the TFHR.

\noindent
\textit{Fuel Parameter Verification}

Verification of fuel parameters, which must be verified with high precision before storing fuel in inaccessible sites, is a major core process in nuclear reactors. Verification is of the correctness of the spent nuclear fuel assembly declarations and the completeness of the declarations associated with the individual nuclear fuel assemblies, with the primary goal of verification being to determine that the spent fuel assembly does indeed contain some nuclear material and that a portion of the assembly has not been diverted \cite{bib:130}. This serves as a major centerpiece of ML applications in reactor physics. Traditional verification of fuel parameters is done through analytical instrumentation. In contrast, based on simulated data and ML methods, Grape et al. \cite{bib:116} systematically explored the ability to predict the initial enrichment, burnup, and cool-down times of fuel parameters independently of operator statements.

\noindent\textit{Signal Anomaly Processing}

To construct an anomaly detection system, it is necessary to be able to capture the fluctuating changes in neutron flux during the operation of the equipment, which involves the ability to identify and analyze the changes in the signal \cite{bib:28}. Caliva et al. \cite{bib:132} proposed a combination of algorithms, including Convolutional Neural Networks (CNNs), Denoising Autoencoders (DAEs), and k-mean clustering, integrated into a system used to analyze neutron flux variations in the core of nuclear reactors. The results show that despite the limited training data and the influence of occlusion or noise on the signals, the perturbation sources can be localized with high accuracy, proving the effectiveness of the method in neutron flux monitoring and anomaly detection in nuclear reactors.

Moreover, Tagaris et al. \cite{bib:28} propose combining wavelet signal analysis with DL techniques. Specifically, they transformed the signal into a scalar map with the help of wavelet transform and used it to train a NN. The results show that the trained network has high accuracy in fault detection and is robust to noise.

\noindent\textit{Digital Twin in Monitoring}

Digital twin technology, an advanced simulation and analysis tool, is increasingly being utilized in the monitoring and maintenance of nuclear reactors. By creating a precise virtual replica of a physical system, digital twins enable the execution of simulation experiments and detailed analyses without disrupting the actual system. This approach significantly enhances the operational efficiency and safety of nuclear reactors by allowing for predictive maintenance, real-time monitoring, and the optimization of system performance under various conditions.

In literature \cite{bib:35}, Chen et al. developed a method to quickly screen and predict the equilibrium burnup state of a nuclear reactor. This method can evaluate the characteristics of the fuel cycle effectively and make fast predictions under different operating conditions. In contrast, Prantikos et al. \cite{bib:39} explored the application of PINN to nuclear reactor digital twin. This approach is not only able to handle data interpolation and extrapolation problems, but also able to make effective predictions in the presence of data scarcity.

In addition, Song et al. \cite{bib:133} proposed a method for the online autonomous calibration of the digital twin. They calibrated the data generated by the digital twin in real-time by combining ML algorithms to match the actual values. The method is implemented through two phases: offline and online, where an error database and a data-driven calibration model are first established in the offline phase, and then dynamically updated and calibrated in the online phase.

Finally, Stewart et al. \cite{bib:134} focused on the development of a real-time extrapolation method to support digital twin solutions for nuclear energy systems. They investigated the ability of the DeepONet method to make real-time predictions of the operating conditions of nuclear reactors without the need for continuous retraining.

\noindent\textit{Fault Monitoring and Diagnostic System}

By leveraging ML methodologies, researchers can monitor and analyze the operational state of nuclear reactors in real time, enabling the early detection of potential faults and anomalies. This proactive approach reduces human error and enhances both the safety and efficiency of reactor operations.

Core monitoring techniques generally encompass methods for detecting core anomalies, characterizing and locating these anomalies, and ultimately classifying them based on their potential impact on the safety and availability of the nuclear power plant. Early identification of such anomalies is crucial for minimizing the risk of accidents. One of the more effective techniques currently available involves measuring neutron noise and analyzing its spatial distribution throughout the reactor core \cite{bib:135}. For instance, Kollias et al. \cite{bib:122} applied ML in noise diagnostics to monitor anomalies using neutron detector measurements. Their approach involved aligning simulated data with real operational data, which enabled the detection of perturbations and facilitated fault classification and location estimation. The latest work by Mena et al. \cite{bib:136} explored the use of autoencoders to detect anomalies in nuclear data that could be potentially used to evaluate the operating status of a nuclear system.

In a study by Nguyen and Diab \cite{bib:109}, a multivariate time series ML model was utilized to predict the transient response of a typical pressurized water reactor (PWR) under a steam generator tube rupture accident (SGTR). By incorporating a Bayesian neural network (BNN), the researchers not only improved the accuracy of the prediction but also quantified the uncertainty of the prediction results.

The Fault Monitoring and Diagnostic Monitoring System (FDDMS) enables real-time monitoring and fault analysis of nuclear reactors in different modes of operation by combining online monitoring (OLM) techniques and data-driven modeling. The core modules of the system include power transient identification, fault detection, and fault diagnosis, each of which employs ML algorithms to process and analyze large amounts of data collected from reactor sensors. The results show that the FDDMS can detect and diagnose faults effectively under different power transient conditions. The system demonstrates the ability to recognize unknown faults and maintains high accuracy in the presence of noise \cite{bib:105, bib:137, bib:138}.

In a study by Nguyen and Diab \cite{bib:109}, a multivariate time series ML model was utilized to predict the transient response of a typical pressurized water reactor (PWR) under a steam generator tube rupture accident (SGTR). By incorporating a BNN, the researchers not only improved the accuracy of the prediction but also quantified the uncertainty of the prediction results, which provides strong support for nuclear safety decision-making.

Zubair and Akram \cite{bib:106}, on the other hand, utilized ML methods in MATLAB to classify transient events in nuclear power plants. By collecting data using the General purpose PWR (GPWR) simulator and combining it with a classification learner application, the researchers successfully identified a wide range of transient events and significantly improved the prediction accuracy and training efficiency of the model through optimization techniques such as feature selection and different validation schemes. They \cite{bib:107} further explored the application of ML techniques in improving the safety and reliability of PWR. By simulating transient events in primary and secondary circuits and using multiple classifiers for training and validation. 

In general, for new nuclear reactors, it may be difficult to strain in case of an accident due to limited experience. Instead, using a virtual nuclear power plant to reproduce behaviors under various conditions and identify unknown anomalies from these behaviors, so that a rapid response can be made to avoid an accident, is a better strategy nowadays \cite{bib:139}. Seki et al. \cite{bib:139} constructed two DNN systems to support the identification of unknown anomalies and determine their causes. Specifically, they estimated the physical quantities of a nuclear power plant in a short time with the help of an agent system, and further applied an anomaly recognition system to estimate the disturbance states that cause anomalies from the physical quantities. They reproduce the steady state, dynamic behavior of the actual High-Temperature Engineering Test Reactor (HTTR) under various scenarios.

\subsubsection{Applications in Safety Design}

Core optimization aims to adjust key state parameters of a reactor core—such as power distribution, temperature field, and reactivity coefficients—to ensure both safe operation and optimal performance. Typically, two major classes of optimization methods are employed: gradient descent–based methods and evolutionary algorithms (EA).

However, because most existing studies rely on external neutron simulation programs\cite{bib:19} to compute these state parameters, the resulting objective functions are often non-differentiable. This non-differentiability can lead to issues like vanishing gradients, which hampers the effectiveness of gradient descent–based optimization.

\noindent\textit{PWR Configuration Optimization}

Back in 2002, in a study by Sadighi et al. \cite{bib:89}, a new approach combining continuous Hopfield Neural Network Artificial (HNNA) and Simulated Annealing (SA) Algorithm was proposed for solving the problem of optimal configuration of fuel assemblies in the core of a pressurized water reactor (PWR). The core of this approach is to utilize the parallel processing capability of neural networks and the global search capability of simulated annealing algorithms to reduce the amount of computation and increase the probability of finding an optimal solution. The study uses the Bushehr Nuclear Power Plant as an example, where neutron flux flattening in the core is used as an objective function to optimize the arrangement of fuel assemblies to achieve a reduction in the local PPF and an increase in the energy production.

To accomplish the optimization of the core configuration in a safe manner, Tayefi et al. \cite{bib:23} used Hopfield neural network artificial (HNNA) to guide a new approach for heuristic search. With the help of a coupled procedure of nuclear code and HNNA, respectively, a cross-section database is built, neutron parameters are calculated, and then the optimal core loading pattern is found by applying the primary fuel assemblies of the VVER/1000 reactor core using the HNNA methodology based on minimizing the PPF and maximizing the $k_\text{eff}$, and the appropriate PPF and $k_\text{eff}$ are obtained. In further study \cite{bib:101}, they also loaded the fuel rods through the proposed fuel rod pattern to find the optimum core configuration by HNNA to find the one that maximizes the $k_{\text{eff}}$.

\noindent\textit{BWR Fuel Array Design}

In the BWR, the design of the nuclear fuel array is critical to ensure safe operation and optimal performance of the reactor. The fuel array consists of a series of fuel rods containing different \(^{235}\text{U}\) enrichments and gadolinium concentrations, which are arranged in a square array, with some positions replaced by water channels \cite{bib:19}. To implement the design, Ortiz et al. \cite{bib:19} demonstrated a system called RENO-CC, which utilizes a multilayer state recurrent neural network (MSRNN) and a fuzzy logic system to optimize the design of the nuclear fuel array at the BWR. With this approach, the researchers were able to minimize the local PPF while keeping the neutron \(k_\text{inf}\) in a given interval, thereby improving the safety and efficiency of the nuclear reactor \cite{bib:19}. Through a series of experiments, RENO-CC successfully designed fuel arrays to meet specific safety and performance requirements and verified their performance with the Core Master PRESTO core simulator.

\noindent\textit{GFR Core Design}

Since reactor designs are often complex, simulations take a considerable amount of time to perform. Kumar et al. \cite{bib:140} developed a genetic algorithm to determine the values of a set of nuclear reactor parameters for the design of a gas-cooled fast breeder reactor (GFR) core, including the underlying thermo-hydraulics analysis and energy transfer. They propose a new method of using regression spline in conjunction with GA, which instead of having to run neutronic simulations on all inputs generated by the GA module, runs simulations on a predefined set of inputs, builds a multiple regression fit to the input and output parameters, and then uses this fit to predict the output parameters of the inputs generated by GA. The reactor core is optimized for high \(^{233}\text{U}\) and \(^{239}\text{Pu}\) multiplication at the peak limit of the required power, required $k_\text{eff}$ and infinite neutron multiplication factor (\(k_\text{inf}\)), high fast fission factor, high thermal efficiency using the Brayton cycle to convert from heat to electricity, and high fuel burnup \cite{bib:140}.

\noindent\textit{Single-channel Design for MSR}

In a molten salt reactor (MSR), the core consists of multiple fuel channels, each containing fuel and coolant, which work together to maintain the critical state of the reactor and to export the heat generated by the reaction. The optimization results of a single-channel design can be used as a starting point for constructing the design of the entire reactor core, which can then be extended to the integrated design and optimization of the entire reactor system \cite{bib:37}. 

To address this problem, Turkmen et al. \cite{bib:37} proposed a robust methodology for the rapid design of nuclear reactor cores and explored the best-performing ML methods for predicting the core's characteristic parameters. Furthermore, they applied the method to a hypothetical molten salt reactor channel to demonstrate the applicability of the method. With the help of an estimator, they found an optimal design for each of the nine fuel salts and estimated all the performance metrics of the design. Among all the fuel salts, the U-Pu-NaCl fuel salt has the highest conversion, the largest negative feedback coefficient, and the lowest fast flux \cite{bib:37}.

\noindent\textit{Core Shape Designs}

Generally speaking, a reactor core is made of industrial materials representing regular component geometries with a periodic structure, that is, one core fuel element is repeated several times to form the entire core \cite{bib:36}. This method produces geometrically similar fuel elements with regular shapes. For further optimization, the literature \cite{bib:36} also considered optimization of core shape from the perspective of shape design.

In the context of fixed-shape core research, Sobes et al. \cite{bib:36} designed a reactor core with geometrically shaped fuel and cooling channels with infinite spatial freedom as elements. They developed an ML-based algorithm for the design and optimization of nuclear reactor cores with flexible geometries and achieved a threefold improvement in the performance metric of peak temperature factor. This research utilized advanced manufacturing techniques, specifically additive manufacturing (3D printing), introduced into nuclear reactor design through the Transformational Challenge Reactor (TCR) program, which makes it possible to design arbitrary geometries for nuclear heating structures. They developed a ML-based multi-physics simulator and evaluated thousands of candidate geometries on Summit, Oak Ridge National Laboratory's leadership supercomputer. By manipulating the core geometry, the results demonstrate the possibility of smoothing the temperature distribution in a nuclear reactor core, which is typically achieved in light water reactors by axial variable component loading and radial fuel swapping \cite{bib:36}.

Addressing the challenges of redundant objectives in shielding design for compact nuclear reactors, Song et al. \cite{bib:110e} introduced a multi-objective shielding optimization method that integrates a deep neural network, principal component analysis, and the NSGA-II algorithm. By employing dimensionality reduction to eliminate redundant objectives, their method successfully balances safety with lightweight design, as demonstrated in both plate shielding models and the Savannah offshore nuclear reactor shielding model.

Looking ahead, it may be beneficial to explore the integration of gradient descent–based methods within deep learning frameworks—potentially through the development of differentiable surrogate models—to further enhance reactor design optimization.

\section{ML-Based Reactor Physics: Future Challenges and Directions}

While ML techniques have made significant strides in reactor physics, it continues to face substantial challenges. The next theoretical or practical advancements remain uncertain. However, we have identified several challenging topics emerging in the literature that highlight key areas of difficulty and suggest potential directions for future research. These challenges serve as a foundation for further exploration and help us contemplate the future trajectory of ML applications in reactor physics.

\subsection{Overcoming Theoretical Difficulties}

In the foundational aspects of reactor physics, as discussed in Section 2, key governing equations and essential state parameters are introduced. The governing equations include the neutron transport equation or neutron diffusion equation, point kinetics equation, and burnup equations. Important state parameters in practical applications include the multiplication factor, reaction cross-section, and neutron flux and power rate distribution, etc. These governing equations form the cornerstone of the deterministic approach in reactor physics. Based on insights from the literature, two primary directions emerge for addressing the theoretical challenges in this field: method-based challenges and model-based challenges. These directions represent the critical avenues for advancing the resolution of complex problems in reactor physics.

\subsubsection{Method-based Challenges}

The frontier ML prediction methodology is represented in reactor physics by DL. The DNN architecture derived from DL is moreover used as a fundamental architecture in many applications.

It is well known that in ML methods, especially supervised or semi-supervised learning, the quality of data has a significant impact on the results. Therefore, one of the challenges lies in the selection of a particular sampling method. In reactor physics, which usually involves a high-dimensional parameter space, traditional sampling methods will become inefficient. During the research process, suitable sampling algorithms can facilitate the training process and provide better approximation \cite{bib:141}, which is one of the future research directions.

As far as a NN is concerned, its core is the selection of the loss function and the structure of the data grid, which can be further investigated in these two aspects. For example, during the research of MSPCNN for seamless encoding of multi-fidelity data \cite{bib:123}, the adaptability and computational efficiency are high, but for this class of networks, it is easy to amplify the error in scenes with limited spatial correlation. Therefore, customizing the loss function to balance the fidelity and reduce the error will be an extension for this problem. In addition, in real-world applications, it is often necessary to model on unstructured or even adaptive grids, where the number and arrangement of grids can be dynamically varied to better capture the phenomena or optimize the computational resources \cite{bib:123}.

Breakthroughs can also be made in terms of the scale of the problem. For larger and more complex problems, the acquisition of high-resolution, high-fidelity data is indeed a significant challenge. The generation of high-fidelity data, such as the statistical Monte Carlo method, is particularly computationally resource intensive. The current mainstream approach consists of fitting high-fidelity data from low-fidelity data with error predictions, e.g., the literature \cite{bib:119} shows how to augment the running simulation of a boiling water reactor (BWR) with the help of ML in combination with traditional simulation techniques. This complementary approach is only in its infancy and is limited to the small reactor problems presented in literature \cite{bib:70}. How to enhance the simulation of high-dimensional problems with the aid of ML for large-scale problems is still under research \cite{bib:7}. It may even be possible to extend this approach to the study of other types of reactors. This is necessary; removing the uncertainty approach to simulation, although for realistic reactors it is possible to obtain operational data, how the data noise is handled should also be taken into account; on top of that, not every demanded parameter has been pre-measured \cite{bib:7, bib:70}.

Finally, the topic can consider the improvement of the algorithm architecture. Specifically, it is the change of the existing network structure. The DL-based PINN method and its variants are particularly used in the review of the main text, which is known for its efficient performance and low mesh dependency. If the Transformer architecture could be integrated into the PINN architecture \cite{bib:123}, or a new auxiliary ML model inspired by Transformer could be created, completely different results could be obtained.

\subsubsection{Model-Based Challenges}

One of important concerns of industry about theoretical models is interpretability. In some cases, reactor operators may be more interested in what input parameters led to the observation or state of interest. Therefore, parameter identification (PI) is another intriguing task \cite{bib:10}. Therefore, the literature \cite{bib:10} has established the digital twin based on non-intrusive modeling and ML methods for the inverse problems for PI, achieving good performance of accurate prediction in a short period. Non-intrusive modeling refers to a modeling approach that does not require modification of the original (forward) model or direct intervention in the physical process. The model can directly utilize the physical model as an input, which retains some physical meaning and improves the interpretability of the model. Another issue is uncertainty quantification \cite{bib:122}. The nuclear reactor itself is a complex system consisting of multi-physics, multi-scale, and for different scales, there may be different parameters, assumptions, and models. To more accurately and comprehensively consider the interpretability of the model, the literature considers multi-scale uncertainty quantification as a future direction of research \cite{bib:129}. Stewart et al. \cite{bib:134} investigated the appropriateness of different ML methods for final implementation by assessing the pros and cons of each method in terms of accuracy, access, development, efficiency, and transfer learning approaches. The authors are also investigating the explainability and interpretability, assess model performance on live data streams, and identify strategies for model verification.

To improve the applicability and reliability of the model, another more critical issue is the generalization of the model, which is also known as the robustness of the model. This is also a solution strategy for multi-scale problems. In most of the literature, the research on ML methods is carried out for specific reactors, how to get rid of the specificity of ML methods for data is a problem to be considered, in other words, a generalization model needs to be found to reduce the dependence on data, or even give a better model without knowing the data in advance \cite{bib:7, bib:48}.

\subsection{Improving Implementation Aspects}

From the perspective of real industrial applications, the addition of a digital twin can remarkably improve the simulation efficiency. The digital twins \cite{bib:10}, as stated by Marguet \cite{bib:15}, as “on-line” agents continuously assimilate models and experimental measurements, which are already known to be faster than real-time processes, should be further investigated. Digital twins could be used along with new data assimilation methods and ML techniques that can be essential in the whole life cycle of a reactor core.

In terms of what has been proposed in literature, the capture of sensor degradation \cite{bib:129} is a research direction. Sensors in nuclear reactors may be degraded by environmental influences or long-term use. ML methods can help to capture the characteristics of the degraded state of the sensors, but the challenge lies in how to accurately identify and process them to ensure the reliability and accuracy of monitoring and controlling the behavior of nuclear reactors.

In addition, strengthening accident handling capabilities with the aid of ML through nuclear reactor dynamics equations \cite{bib:142} can effectively enhance the safety of nuclear reactors. Nuclear reactor accident handling is a complex process, which needs to consider a variety of factors and uncertainties, how to quantify the uncertainty well, how to enhance the anomaly monitoring capability, how to improve the prediction performance, and so on, should also be investigated.

Symbolic language conversion is a relatively new topic \cite{bib:122}. ML methods usually use mathematical models and ML algorithms for modeling and analysis, which makes it difficult for humans to understand the model results. To integrate theory with practice, research is needed on how to transform the results of ML models into a human-understandable symbolic language that can better support decisions and applications.

Finally, there is the challenge of making accurate noise observations of measurement data \cite{bib:7, bib:136}. Accurate observation of noise is critical for model prediction and system monitoring. ML methods can help to improve the accuracy of noise observations, but the challenge of how to efficiently process noisy data and differentiate between real signals and noise needs to be addressed.

\section{Conclusions}

This review provides an in-depth study of the research directions and research content of the application of ML methods to nuclear reactor physics since their development. At present, the application of ML methods in reactor physics focuses on the theoretical and applied directions. On one hand, the research begins from the efficiency of solving governing equations and the efficiency of predicting reaction state parameters, the former with the help of DL technology, in which the PINN algorithm and its variants have yielded excellent results. The latter, on the other hand, focuses on the multiplication factor, neutron cross-section, and neutron flux, and mainly focuses on reducing the amount of computation while maintaining the accuracy of the solution, i.e., an innovative approach that unifies efficiency and accuracy. In terms of the solution of nuclear reactor problems based on the ML methods, it can be seen as a response to the equations and parameters in Section 2 on how to solve practical problems. Some of the difficult problems in reactor physics can be developed in three important areas: the steady-state, the transient state, and the burnup. Steady-state problems consider the prediction of performance, temperature field, and core power, including the estimation of the power peaking factor, etc., and provide advancement mainly for areas such as core design. The transient problem is concerned with reactor behavior monitoring and safety analysis, especially how to respond quickly and make safe decisions under accident conditions. As for burnup, it is considered from the perspective of fuel composition, burnup distribution, and other reaction environments. Furthermore, advancements in nuclear engineering are increasingly driven by simulation, regulation, and design of nuclear reactors, effectively bridging the theoretical challenges discussed in the previous chapter with practical applications. ML methods play a crucial role in data generation, anomaly identification, and noise management. Additionally, concepts such as data assimilation, agent-based modeling, and digital twin technology are being incorporated into practical applications, serving as prime examples of how ML methods can be seamlessly integrated with complex nuclear systems. These developments highlight the potential of ML to enhance both the safety and efficiency of nuclear reactor operations. 

In conclusion, the application of ML methods in nuclear reactor physics holds significant potential for further development. Theoretically, there is room for enhancing ML methods and structures, particularly in improving the interpretability and generalization of models. From a broader perspective, challenges remain in areas such as optimizing simulation performance, accelerating anomaly detection, refining accident handling, and enhancing safety design in industrial settings. Additionally, the industry faces substantial challenges, including the need for objective exclusion, the expansion of digital twin technology, and the pursuit of better data parsimony. Furthermore, it should be emphasized that machine learning in the coupling of reactor physics and thermo-hydraulics is still an important topic that needs more research and review. Addressing these challenges will be crucial for advancing the integration of ML in nuclear reactor physics and engineering.

\section*{Acknowledgments}
This research was partially sponsored by the Natural Science Foundation of Shanghai (no. 23ZR1429300), the Innovation Funds of CNNC (Lingchuang Fund, contract no. CNNC-LCKY-202234), and the National Natural Science Foundation of China (grant no.12471377).

\end{document}